\documentclass[Afour,sageh,times]{sagej}

\usepackage{moreverb,url}
\usepackage[colorlinks,bookmarksopen,bookmarksnumbered,citecolor=red,urlcolor=red]{hyperref}

\newcommand\BibTeX{{\rmfamily B\kern-.05em \textsc{i\kern-.025em b}\kern-.08em
T\kern-.1667em\lower.7ex\hbox{E}\kern-.125emX}}

\usepackage{natbib}
\usepackage{siunitx}
\usepackage{subcaption}
\usepackage[section]{placeins}
\usepackage{fancyhdr}
\usepackage{mathtools}
\usepackage{amssymb}
\usepackage{amsmath}
\usepackage{dblfloatfix} 
\usepackage{comment}
\usepackage{hyperref}
\usepackage{graphicx}
\usepackage{subcaption}
\usepackage{caption}
\usepackage{xcolor}
\usepackage{siunitx}
\usepackage{booktabs}
\usepackage{nicefrac}
\usepackage[ruled,vlined,linesnumbered]{algorithm2e}
\usepackage[normalem]{ulem}

\newcommand{\norm}[1]{\left\lVert#1\right\rVert}
\newcommand{\ie}{{i.e. }}

\newcommand{\realR}{\mathbb{R}}

\usepackage{soul}
\makeatletter
\renewcommand\sout[1]{\@bsphack\@esphack}
\makeatother

\def\volumeyear{2023}

\begin{document}

\runninghead{Ryou et al.}

\title{Multi-Fidelity Reinforcement Learning for Time-Optimal Quadrotor Re-planning}

\author{Gilhyun Ryou, Geoffrey Wang, and Sertac Karaman}

\affiliation{Laboratory for Information and Decision Systems, Massachusetts Institute of Technology, Cambridge, MA, USA}

\corrauth{Gilhyun Ryou, Laboratory for Information and Decision Systems, \\
Massachusetts Institute of Technology, 77 Massachusetts Avenue, \\
Cambridge, MA 02139, USA.}

\email{ghryou@mit.edu}

\begin{abstract}
High-speed online trajectory planning for UAVs poses a significant challenge due to the need for precise modeling of complex dynamics while also being constrained by computational limitations. 
This paper presents a multi-fidelity reinforcement learning method (MFRL) that aims to effectively create a realistic dynamics model and simultaneously train a planning policy that can be readily deployed in real-time applications.
The proposed method involves the co-training of a planning policy and a reward estimator; the latter predicts the performance of the policy's output and is trained efficiently through multi-fidelity Bayesian optimization. 
This optimization approach models the correlation between different fidelity levels, thereby constructing a high-fidelity model based on a low-fidelity foundation, which enables the accurate development of the reward model with limited high-fidelity experiments.
The framework is further extended to include real-world flight experiments in reinforcement learning training, allowing the reward model to precisely reflect real-world constraints and broadening the policy's applicability to real-world scenarios.
We present rigorous evaluations by training and testing the planning policy in both simulated and real-world environments.
The resulting trained policy not only generates faster and more reliable trajectories compared to the baseline snap minimization method, but it also achieves trajectory updates in 2 ms on average, while the baseline method takes several minutes.
\end{abstract}


\maketitle
 
\section*{Supplementary Material}
% A video and code of the experiments are available at our project website:
% \url{https://aera.mit.edu/projects/xxx}.
The experimental results are included in the following video:
\url{https://youtu.be/75AbKY3L5As}.
Detailed implementation and source code are available at:
\url{https://github.com/mit-aera/mfrlTrajectory}.

% !TeX root = ../root.tex

\section{Introduction} \label{sections:introduction}

Traditional high-performance motion-planning algorithms typically rely on mathematical models that capture the system's geometry, dynamics, and computation.
Model predictive control, for example, utilizes these models for online optimization to achieve desired outcomes. 
However, due to computational constraints and the complexity of the inherent non-linear optimization, these approaches often depend on simplified models, leading to conservative solutions. 
Alternatively, data-driven approaches like Bayesian Optimization (BO) and Reinforcement Learning (RL) offer more complex modeling for better planning solutions. 
BO is particularly effective at identifying optimal hyperparameters from a limited dataset, beneficial for modeling real-world constraints. 
Yet, its real-time application is limited by the need to estimate model uncertainty.
RL, on the other hand, constructs high-dimensional decision-making models representing intricate behaviors and is conducive to GPU acceleration for real-time use. 
However, given  its substantial need for training data, it is often trained in simulated environments and later adapted to real-world scenarios through Sim2Real techniques, a process that might compromise performance for adaptability.

\begin{figure}[thb]
\centering
\includegraphics[width=0.48\textwidth,trim=.0cm 0.cm .0cm .0cm,clip]{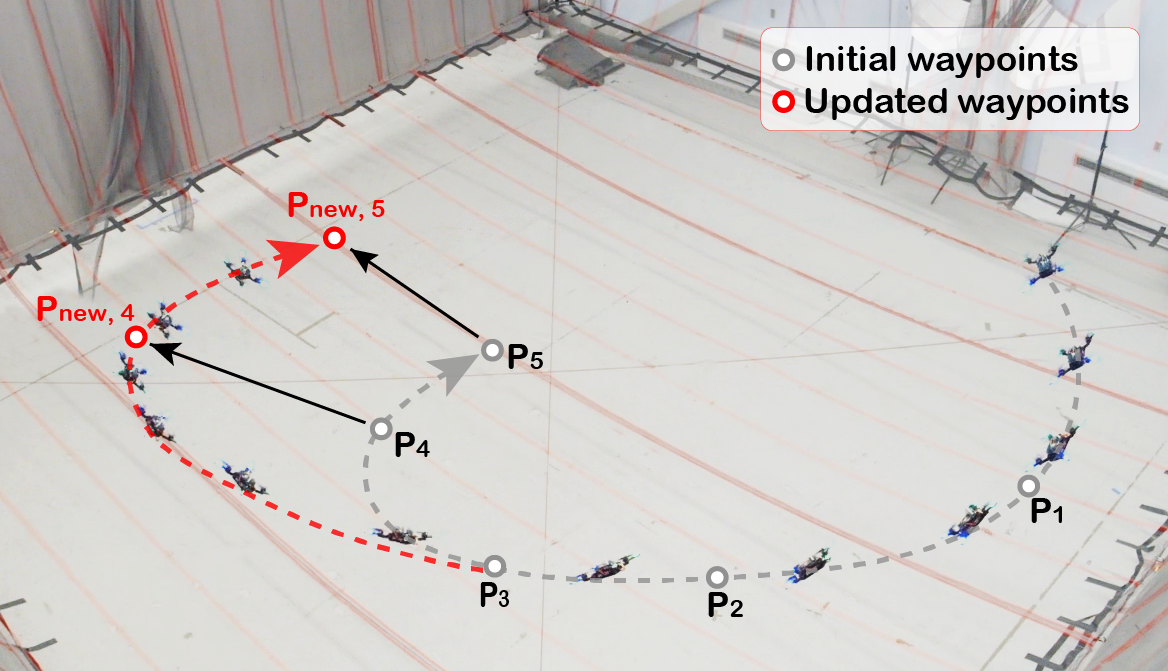}
\vspace{-0.5em}
\caption{Time-optimal re-planning problem: While a quadrotor vehicle passes through the waypoints, the remaining waypoints are randomly shifted, necessitating real-time trajectory adaptation.}
\label{fig:intro_main}
\vspace{-1em}
\end{figure}

% This paper addresses the challenge of time-optimal online trajectory planning for agile vehicles, a vital component in various unmanned aerial vehicle (UAV) applications.
This paper addresses the challenge of time-optimal online trajectory planning for agile vehicles with dynamic waypoint changes, as illustrated in Figure \ref{fig:intro_main}.
This capability represents a vital component in various unmanned aerial vehicle (UAV) applications.
However, crafting high-speed quadrotor trajectories is particularly demanding due to the complex aerodynamics, turning trajectory generation into a time-consuming non-linear optimization. 
Moreover, real-world constraints, including the effects of control delay and state estimation error, can make the generated trajectory unreliable even if it satisfies the ideal dynamics constraints. 
Online planning adds yet another layer of complexity by incorporating computational time constraints to update the trajectory in real-time. 
As a consequence, many existing online planning methods rely on simplified constraints such as bounding velocity and acceleration~\cite{gao2018online, tordesillas2019faster, wu2023learning}, or on hierarchical planning structures that combine the approximated global planner with the underlying control schemes~\cite{romero2022time}, which may lead to conservative solutions.

The main contribution of this paper is a novel multi-fidelity reinforcement learning (MFRL) framework. 
MFRL integrates RL and BO to develop a planning policy optimized for real-world and real-time scenarios.
First, this method utilizes BO to directly model feasibility boundaries from high-fidelity evaluations—accurate but expensive methods, e.g., real-world experiments. 
It improves policy performance by accurately modeling constraint boundaries where time-optimal trajectories are typically found. 
Second, we implement multi-fidelity Bayesian optimization (MFBO) to reduce required high-fidelity evaluations by using low-fidelity evaluations—quick but less accurate methods, e.g., analytical checks with ideal dynamics—to rapidly screen out excessively fast or slow trajectories with obvious tracking results. 
Thirdly, RL uses this constraint modeling in the reward signal for policy training, resulting in a computationally efficient policy that enables real-time generation of time-optimal trajectories for online planning scenarios.
Lastly, the trained policy is validated through comprehensive numerical simulations and real-world flight experiments. 
It outperforms the baseline minimum-snap method by achieving up to a 25 \% reduction in flight time, averaging a 4.7 \% decrease, with just 2 ms of computational time—markedly less than the several minutes required by the baseline.

% !TeX root = ../root.tex

\section{Related Work} \label{sections:related_works}

\subsection{Quadrotor Trajectory Planning}
Quadrotor planning often employs snap minimization, leveraging the differential flatness of quadrotor dynamics for smooth trajectory generation~\cite{mellinger2011minimum, richter2016polynomial}.
Minimizing the fourth-order derivative of position, \ie, snap, and the yaw acceleration produces a smooth trajectory that is less likely to violate feasibility constraints. 
Several methods that outperform the snap minimization method have been presented in recent years.
\citet{foehn2021time} employs a general time-discretized trajectory representation, using waypoint proximity constraints to find better, time-optimal solutions.
\citet{gao2018optimal} streamlines minimum-snap computation by partitioning the optimization problem, separating speed profile and trajectory shape optimizations. 
% \citet{sun2021fast} employs spatial optimization gradients to solve non-linear optimization for time allocation, which offers numerical stability advantages over naive non-linear optimization. 
% \citet{burke2020generating} further improves the numerical stability of the spatial optimization by reformulating quadratic programming with matrix factorization.
\cite{sun2021fast, burke2020generating} employ spatial optimization gradients to solve non-linear optimization for time allocation, which offers numerical stability advantages over naive non-linear optimization. 
\citet{qin2023time} improves the time optimization by incorporating linear collision avoidance constraints into the objective function through projection methods, transforming the problem into unconstrained optimization. 
\citet{wu2023learning} utilizes supervised learning to learn time allocation, minimizing the computation time for time optimization. 
These approaches optimize trajectory time while adhering to the snap minimization objective. 
\citet{mao2023toppquad}, however, directly minimizes time using reachability analysis within speed and acceleration bounds.
Instead of using the predefined velocity and acceleration constraints, \citet{ryou_tal_ijrr} utilizes MFBO to model the feasibility constraints from the data and utilizes the model to directly minimize trajectory time.

Most offline trajectory generation methods, due to the limited computation time, require approximation to be extended to the online replanning problem.
\citet{gao2018online} formulates the online planning problem as quadratic programming that can be solved in real-time by directly optimizing the velocity and acceleration that conforms to the specified velocity and acceleration constraints.
\citet{tordesillas2019faster} uses the time allocations from the previous trajectory to warm-start the optimization for the updated trajectory, thereby reducing the computation time.
% Explain heuristics
\citet{romero2022time} builds a velocity search graph between the prescribed waypoints sequence and finds the optimal velocity profile with Dijkstra search.
As the trajectory is generated in discrete state space, the resulting trajectory may not be feasible, so feasible control inputs are generated using underlying model predictive control~\cite{romero2022model} or deep neural network policy~\cite{molchanov2019sim}.
On the other hand, \citet{kaufmann2023champion} trains a neural network policy to directly generate thrust and body rate commands based on the vehicle's position using RL, achieving superior performance even surpassing human racers. Nonetheless, further research is needed to expand the applicability of this learning-based approach to deploy the trained policy effectively in unseen, diverse environments.

These online planning methods often employ local planning with discrete state space optimization, allowing for greater expressiveness and easier implementation of constraints.
In contrast, our paper opts for a polynomial trajectory representation.
We choose this representation for its reduced optimization dimension, facilitating lighter neural networks and easier dataset creation for machine learning. 
Additionally, this approach allows for longer planning horizons, incorporating more future information for better overall performance. 
While the polynomial representation has trade-offs compared to discrete approaches, it has potential to improve expressiveness through additional variables like knot points or higher-degree polynomials.

\subsection{Efficient Reinforcement Learning}
The sampling efficiency must be improved in order to use RL in practical settings where computation time must be kept manageable.
% Model-based RL
The model-based RL utilizes a predefined or a partially learned transition model to guide the policy search towards the feasible action space.
\citet{sutton1990integrated} trains the model of the transition function and then uses this model to generate extra training samples.
\citet{deisenroth2011pilco} employs Gaussian processes to approximate the transition function, which allows model-based RL to be applied to more complex problems.
Similarly, \citet{gal2016improving} extends the same framework by utilizing a Bayesian neural network as the transition function model.
Model-based RL is widely used in the robotics applications to train controllers or design model-predictive control schemes~\citep{williams2017information, nguyen2010using, xie2018feedback}.
Among these robotics applications, \citet{cutler2015real} uses model-based learning with multi-fidelity optimization by merging data from simulations and experiments, which is relevant to the proposed method.

% Uncertainty aware reinforcement learning
Uncertainty-aware RL is an additional strategy to boost sample efficiency, which uses uncertainty estimation to streamline exploration throughout the RL training phase.
\citet{osband2016deep} uses bootstrapping method to estimate policy output variance from dataset subsets, enhancing sample efficiency by selecting outputs that maximize rewards from the policy outputs' posterior distribution.
\citet{kumar2019stabilizing} uses the same method of uncertainty estimation to evaluate out-of-distribution samples, aiming to stabilize the training process. 
\citet{wu2021uncertainty} utilizes MC-dropout, which is introduced in \cite{gal2016dropout}, to predict the variance of predictions and regularizes the training error with the inverse of this variance. 
Similarly, \citet{clements2019estimating, lee2021sunrise} apply an ensemble method, which involves using multiple policy models to estimate the uncertainty, facilitating safer exploration during RL.
Reviews of these methodologies are included in \citet{lockwood2022review, hao2023exploration}.
To handle high-fidelity samples with very limited data, we adopt a Gaussian process framework with a multi-fidelity kernel, moving away from neural-network-based uncertainty estimation. 
We instead compensate for the limited model capacity of the Gaussian process using a variational approach, elaborated on in the subsequent section.

% Reward learning
Using a separately learned reward model also can improve the sample efficiency of RL.
% These method utilizes a simple model, such as linear feature-based regression or Gaussian processes, to learn 
% The policy model to maximize the trained reward function, and as a result, these methods can not only handle a non-numerical or intrinsic reward function, but they can also improve training efficiency.
For instance, preference-based RL is frequently used when dealing with an unknown reward function that is difficult to design even with expert knowledge~\citep{abbeel2010autonomous, wirth2017survey, biyik2020active}.
These methods often utilize a simple model, such as linear feature-based regression or Gaussian process~\citep{biyik2020active}, to learn the reward model from a small number of expert demonstrations.
% The policy model to maximize the trained reward function, and as a result, these methods can not only handle a non-numerical or intrinsic reward function, but they can also improve training efficiency.
\citet{konyushkova2020semi} utilizes the idea of efficient reward learning in offline learning problem, and trains the policy in semi-supervised manner.
Similar approaches are used in safety-critical systems where the number of real-world experiments must be kept to a minimum.
For example, \citet{srinivasan2020learning} trains the model of feasibility function and use it to estimate the expected reward in order to safely apply RL.
\citet{zhou2022dynamic} utilizes generative adversarial network (GAN) loss to train the model in a self-supervised manner.
\citet{christiano2017deep} updates the reward model by incorporating human feedback, which is provided in the form of binary preferences for policy outputs. 
Likewise, \citet{glaese2022improving} employs a human feedback-based approach to train large language models using RL.

% Sim2Real
Several methods have been proposed to deploy trained models in real-world robotic systems, addressing the discrepancies between simulated and real-world environments. 
One approach involves introducing random perturbations to the input of policy (\cite{sadeghi2018sim2real, akkaya2019solving}) or dynamics model (\cite{peng2018sim, mordatch2015ensemble}) during RL training. 
This method produces robust policies capable of handling errors related to the simulation-to-reality gap.
However, this may compromise the policy's performance, making it unsuitable for tasks that demand precise modeling of the system's limits, like time-optimal trajectory generation.
Alternatively, the real-world deployment can be enhanced by increasing the accuracy of the simulation.
For instance, using system identification techniques, studies such as \cite{tan2018sim, kaspar2020sim2real, hwangbo2019learning} incorporate real-world constraints like actuation delay and friction into simulations.
Meanwhile, \cite{chang2020sim2real2sim, lim2022real2sim2real, chebotar2019closing} iteratively refine simulations based on real-world data using learning-based methods. 
These approaches improve simulation accuracy and computational efficiency, making them more suitable for RL training. 
However, some components, like battery dynamics, remain challenging to model accurately. 
Moreover, complex systems such as autonomous vehicles often can't be comprehensively simulated. 
These limitations necessitate real-world experiments on robot deployment, which our paper aims to make more efficient.

\subsection{Bayesian Optimization}
In this work, we utilize Bayesian optimization techniques (BO) to improve the efficiency of real-world experiments. 
These methods are particularly useful in fields where the efficient use of limited experimental resources is crucial, such as in pharmacology (\cite{lyu2019ultra}), analog circuit design (\cite{zhang2019efficient}), aircraft wing design (\cite{rajnarayan2009trading}), physics (\cite{dushenko2020sequential}), and psychology (\cite{myung2013tutorial}). 
BO continuously fine-tunes a surrogate model that estimates the information gain and uses this model to strategically choose subsequent experiments. 
Utilizing data from prior experiments, it creates a surrogate model to predict the information gain of future experimental options. 
BO then selects experiments from the available choices that best reduce uncertainty in the surrogate model (\cite{takeno2019multi, costabal2019multi}) and precisely locates the optimum of the objective function (\cite{mockus1978application, hernandez2014predictive, wang2017max, srinivas2012information}).
For the surrogate model, the Gaussian Process (GP) is often chosen due to its efficiency in estimating uncertainty from a limited number of data points (\cite{williams2006gaussian}). 
To extend GP's applicability to large, high-dimensional datasets, several scalable methods have been developed. 
The inducing points method is particularly notable for big datasets, employing pseudo-data points for uncertainty quantification (\cite{snelson2006sparse, hensman2013gaussian}) instead of the full data set. 
Additionally, the deep kernel method (\cite{wilson2016deep, calandra2016manifold}) is utilized for high-dimensional data, compressing it into low-dimensional feature vectors via deep neural networks before applying GP.

When multiple information sources are available, multi-fidelity Bayesian optimization is employed to merge information from different fidelity levels. 
This approach leverages cost-effective low-fidelity evaluations, like basic simulations or expert opinions, to refine the design of more expensive high-fidelity measurements such as complex simulations or real-world experiments.
To incorporate information from multiple sources, the surrogate model must be modified to combine multi-fidelity evaluations.
\cite{kennedy2000predicting, le2014recursive} apply a linear transformation between Gaussian Process (GP) kernels to model the relationship between different fidelity levels. 
\cite{perdikaris2017nonlinear,cutajar2019deep} extend this method by including a more sophisticated nonlinear space-dependent transformation.
Additionally, \citet{dribusch2010multifidelity} utilizes the decision boundary of a support vector machine (SVM) to reduce the search space of high-fidelity data points, and \citet{xu2017noise} utilizes the pairwise comparison of low-fidelity evaluations to determine the adversarial boundary of the high-fidelity model. 

%%%%%%%%%%%%%%%%%%%%%%%%%%%%%%%%%%%%%%%%%%%%%%%%%%%%%

\section{Preliminaries} \label{sections:preliminaries}

\subsection{Minimum-Snap Trajectory Planning}
% Quadrotor planning often employs snap minimization, leveraging the differential flatness of quadrotor dynamics for smooth trajectory generation~\cite{mellinger2011minimum, richter2016polynomial}.
% Minimizing the fourth-order derivative of position, \ie, snap, and the yaw acceleration produces a smooth trajectory that is less likely to violate dynamics and real-world constraints. 
The snap minimization method utilizes the smooth polynomial that is obtained by minimizing the fourth-order derivative of position and the second-order derivative of yaw.
For a sequence of waypoints $\tilde p^i$, each consisting of a prescribed position $\tilde p^i_r \in \mathbb{R}^3$ and yaw angle $\tilde p^i_\psi \in \mathbb{S}^1$ ($\tilde{\mathbf{ p}} = \begin{bmatrix} \tilde p^0 \cdots \tilde p^m \end{bmatrix}$), this approach models the trajectory using piecewise continuous polynomials. 
These polynomials map time to position and yaw, \ie, $p(t) = \begin{bmatrix} {{}p_r(t)}^\intercal &  p_\psi(t) \end{bmatrix}^\intercal$ ($p_r(t) \in \mathcal{C}^4, p_\psi(t) \in \mathcal{C}^2$), and each polynomial segment connects two adjacent waypoints.
$\mathbb{S}^1$ denotes the circular angle space, and $\mathcal{C}^n$ is the differentiability class where its n-th order derivatives exist and are continous.

The coefficients of the polynomial trajectory are determined by solving the following optimization problem:
\begin{equation}
\begin{gathered}
\underset{p=[p_r, p_\psi], \mathbf{x} \in \realR^m_{\geq 0}}{\text{minimize}} \;\;\; \sigma\left(\mathbf{x}, p\right) + \rho {\sum \nolimits}_{i=1}^{m}x_i\\ 
\text{subject to}\;\;  p(0) = \tilde{p}^0,\;\; p\left({\textstyle\sum \nolimits}_{j=1}^{i}x_j\right) = \tilde{p}^i, \; i=1,\;\dots,\;m, \\
\;\;\;\; p \in \mathcal{P}
\label{eqn:naiveminsnap}
\end{gathered}
\end{equation}
where 
\begin{equation}\label{eqn:smoothness}
\sigma(\mathbf{x}, p)=\int_{0}^{{\textstyle\sum \nolimits}_{i=1}^{m}x_i} \mu_r \norm{\frac{d^4 p_{r}}{d^4 t}}^2 + \mu_\psi \Big(\frac{d^2 p_\psi}{d^2 t}\Big)^2 dt
\end{equation}
with $\mu_r$, $\mu_\psi$ and $\rho$ are weighting parameters, and $\mathcal{P}$ represents the set of feasible trajectories.
$x_i$, the pivotal optimization variable, denotes the time allocation between two consecutive waypoints $\tilde p_{i-1}$ and $\tilde p_i$ ($\mathbf{x} = \begin{bmatrix} x_1 \cdots x_m \end{bmatrix}$), and is exclusively determined through non-linear optimization.

% Shrink this
This method consists of three components to efficiently solve the optimization problem: 
1) Inner-loop spatial optimization that uses quadratic programming to derive polynomial coefficients based on the time allocation between waypoints:
\begin{equation}
\begin{gathered}
\chi(\mathbf{x}, \tilde{\mathbf{ p}}) = \underset{p=[p_r, p_\psi]}{\text{argmin}} \;\;\; \sigma\left(\mathbf{x}, p\right) \\
\text{subject to}\;\;  p(0) = \tilde{p}^0,\;\; p\left({\textstyle\sum \nolimits}_{j=1}^{i}x_j\right) = \tilde{p}^i, \; i=1,\;\dots,\;m,
\end{gathered}
\end{equation}
2) Outer-loop temporal optimization via non-linear programming that minimizes the output of the inner-loop quadratic programming and determines the time allocation ratio:
\begin{equation}\label{eqn:minsnap_nonlinear}
\underset{\tilde{\mathbf{x}} \in \realR^m_{\geq 0}}{\text{minimize}} \;\;\; \sigma\left(\chi(\tilde{\mathbf{x}}, \tilde{\mathbf{ p}}), \tilde{\mathbf{x}}\right) \;\;
\text{subject to}\;\;  {\sum \nolimits}_{i=1}^{m}\tilde{x}_i = 1,
\end{equation}
3) Line search to generate a feasible trajectory by scaling the time allocation ratio acquired from the outer-loop optimization:
\begin{equation}\label{eqn:minsnap_linesearch}
  \underset{\alpha \in \realR_{> 0}}{\text{minimize}}\;\alpha,\;\;\text{subject to} \;\;\chi(\alpha \tilde{\mathbf{x}}, \tilde{\mathbf{ p}}) \in \mathcal{P}
\end{equation}

When a vehicle starts and ends in a stationary state, with zero velocity and acceleration, uniformly scaling time allocations does not alter the trajectory's shape but shifts control commands away from the default stationary control commands. 
Consequently, the snap minimization method can determine feasible trajectory timings by scaling the time allocations after optimizing the time allocation ratio using quadratic programming and nonlinear optimization. 
Figure \ref{fig:alg_eval_compare_ms} shows how the line search procedure changes the speed profile and motor speed commands.
However, this approach is not effective for re-planning from non-stationary states, as scaling then alters the trajectory shape and control command topology. 
The feasibility constraints, $\mathcal{P}$, can encompass various conditions, such as trajectories with admissible control commands or those with tracking errors within a certain threshold. 
Yet, due to computational constraints in real-time applications, these are often simplified to velocity and acceleration limits. 
This article primarily aims to train a planning policy capable of directly outputting time allocations from non-stationary states and applying more precise feasibility constraints for policy training.

\subsection{Multi-Fidelity Gaussian Process}
% Add how many samples are used for the training?
In this paper, we employ the Multi-fidelity Gaussian Process Classifier (MFGPC) to efficiently model the feasibility boundary based on sparse multi-fidelity evaluations. 
Given a collection of data points $\mathbf{Z}=\{\mathbf{z}_1,\cdots,\mathbf{z}_N\}$ with their associated evaluation outcomes from the $l$-th fidelity level $\mathbf{y}_l=\{y_{l,1},\cdots,y_{l,N}\}$, the prior Gaussian distribution is modeled using a multi-fidelity covariance kernel.
The key idea of the multi-fidelity covariance kernel $K_l(\mathbf{Z},\mathbf{Z})$ is utilizing cheap low-fidelity evaluations to efficiently model the high-fidelity measurements.
The relationship between the distributions of different fidelity levels is represented by the nonlinear space-dependent transformation proposed in \citet{cutajar2019deep}.
To be specific, $k_l(z,z')$, an element of the $l$-th covariance kernel $K_l(\mathbf{Z},\mathbf{Z})$ corresponding to data points $z$ and $z'$, is estimated as
\begin{equation}
\begin{aligned}
    k_l(z,z') = k_{l,corr}(z,z')(\sigma_{l,linear} ^2 \:f_{l-1}(z)^T f_{l-1}(z') \\
    + k_{l,prev}(f_{l-1}(z), f_{l-1}(z'))) + k_{l,bias}(z,z'),
\end{aligned}
\label{eqn:alg_mfdgp_covariance}
\end{equation}
where $f_{l-1}$ is the Gaussian process estimation from the preceding fidelity level, $\sigma_{l,linear}$ is a constant scaling the linear covariance kernel, and 
$k_{l,prev}$, $k_{l,corr}$ and $k_{l,bias}$ represent the covariance with the preceding fidelity, the space-dependent correlation function and the bias function, respectively.

\begin{figure}[!h]
\centering
\includegraphics[width=0.4\textwidth,trim=0.cm 0.cm 0.cm 0.cm,clip]{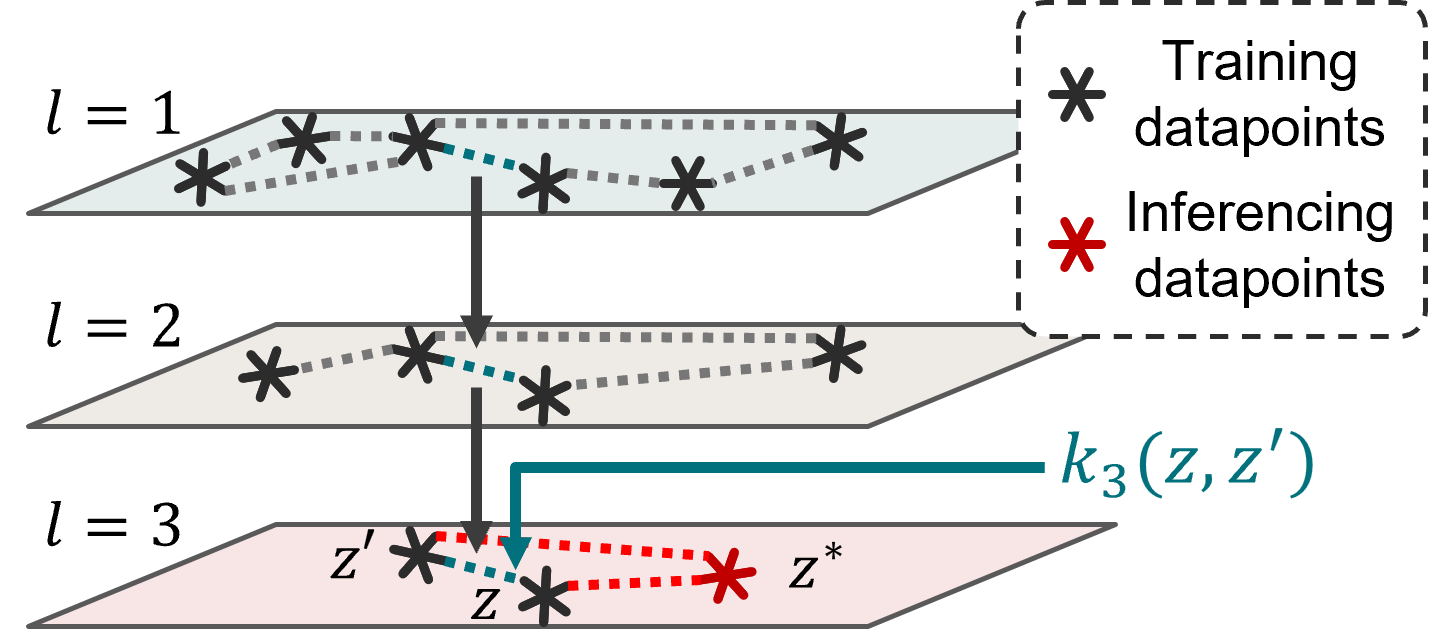}
% \vspace{-.5em}
\caption{
Multi-fidelity Gaussian process classifier training and inferencing. 
Training: Lower fidelity kernel captures the overall structure using ample data to optimize more hyperparameters. 
Higher fidelity learns linear transformation on top. 
Covariance between dataset pairs, $z$ and $z'$, (Cyan line) is estimated using previous level guidance. 
Inferencing: Covariances estimated between existing and a new data $z^*$ (Red line) are used for prediction via marginalization of training variables.
}
\label{fig:rel_mfgp}
\end{figure}

The covariance kernel is used to model a joint Gaussian distribution of the evaluations and the latent variables $\mathbf{f}=\begin{bmatrix}f_1,\cdots,f_{N}\end{bmatrix}$. 
The latent variables and the hyperparameters of the kernel function are trained by maximizing the marginal likelihood function
\begin{align}
P(\mathbf{y},\mathbf{f}|\mathbf{Z})&=\Pi_{i=1}^N P(y_{i}|f_{i})P(\mathbf{f}|\mathbf{Z})\;\\
&=\Pi_{i=1}^N \mathcal{B}(y_{i}|\Phi(f_{i}))\mathcal{N}(\mathbf{f}|0,K_l(\mathbf{Z},\mathbf{Z})),
\end{align}
where $\mathcal{B}(x)$ is the Bernoulli likelihood and $\Phi(f_{l,i})$ is the cumulative density function used to map the latent variable $f_{l,i}$ onto the probability domain [0, 1].
The prediction, $y_*$, on the l-th fidelity level over a new data point, $z_*$, is obtained by marginalizing the latent variable, $f_*$, as follows:
\begin{equation}
    P(y_*|\mathbf{z}_*,\mathbf{Z},\mathbf{y}) = \int P(y_*|f_*)P(f_*|\mathbf{z}_*,\mathbf{Z},\mathbf{y}_l)d\mathbf{f_*}.
\end{equation}
Figure \ref{fig:rel_mfgp} illustrates the training and inferencing procedure of the multi-fidelity Gaussian process classifier.

MFGPC is further accelerated with the inducing points method~\cite{snelson2006sparse, hensman2013gaussian, hensman2015scalable}, which approximates the distribution $P(\mathbf{f}|\mathbf{Z})$ by introducing a variational distribution $q(\mathbf{u})=\mathcal{N}(\mathbf{m},\mathbf{S})$.
The hyperparameters $\mathbf{m},\mathbf{S}$ represent the mean and covariance of the inducing points $\mathbf{u}$.
The method minimizes the following variational lower bound as loss function to determine the inducing points and maximizes the marginal likelihood $P(\mathbf{y},\mathbf{f})$:
% \begin{equation}
% \mathcal{L}(\mathbf{y},\mathbf{Z}) = -\sum_{i=1}^N \mathbb{E}_{q_i(f_i)} \left[ \log P(y_i|f_i) \right] + D_{\text{KL}} \left[ q(\mathbf u) \Vert P(\mathbf u|\mathbf{Z}) \right],
% \label{eqn:alg_modeling_mfdgp_loss}
% \end{equation}
\begin{equation}
\mathcal{L}_{\mathcal{M}}(\mathbf{y},\mathbf{Z}) = -\mathbb{E}_{q(\mathbf{f})} \left[ \log P(\mathbf{y}|\mathbf{f}) \right] + D_{\text{KL}} \left[ q(\mathbf u) \Vert P(\mathbf u|\mathbf{Z}) \right],
\label{eqn:alg_modeling_mfdgp_loss}
\end{equation}
% \begin{equation}
% \begin{aligned}
% \mathcal{L}(\mathbf{y},\mathbf{f},\mathbf{Z}) = &-{\textstyle\sum \nolimits}_{i=1}^N \mathbb{E}_{q_i(f_i)} \left[ \log P(y_i|f_i) \right] \\
% &\;\;\; + D_{\text{KL}} \left[ q(\mathbf u) \Vert P(\mathbf u|\mathbf{Z}) \right],
% \label{eqn:alg_modeling_mfdgp_loss}
% \end{aligned}
% \end{equation}
where $q(\mathbf{f}) \coloneqq \int P(\mathbf{f}|\mathbf{u})q(\mathbf{u})d\mathbf{u}$. 
$\mathcal{M} \coloneqq \left(\mathbf{f}, \theta, \mathbf{m}, \mathbf{S}\right)$ is the set of all hyperparameters, and $\theta$ is the hyperparameters of the Gaussian process kernel.
The hyperparameters of the MFGPC are determined by minimizing the sum of the variational lower bounds across all fidelity levels:
\begin{equation}
\mathcal{L}_{\text{MFGPC}} = \sum_{l=1}^{L} \left(\mathcal{L}_{\mathcal{M}_l}(\mathbf{y}_l,\mathbf{Z}_l) + c_{\text{reg}}\sum_{l^\prime=1}^{l-1} \mathcal{L}_{\mathcal{M}_{l^\prime}}(\mathbf{y}_l,\mathbf{Z}_l)\right).
\label{eqn:alg_modeling_mfdgp_loss_all}
\end{equation}
The losses incurred at lower fidelity levels, $\mathcal{L}_{\mathcal{M}_{l^\prime}}$ where $l' \in \{1, \dots, l-1\}$, are added to provide regularization, with $c_{\text{reg}}$ representing the coefficient for the regularization term.
During the MFRL training, samples with high uncertainty estimates from the boundary model are selected, enabling iterative multi-fidelity evaluations and boundary model updates.
MFGPC is implemented with GPyTorch~\cite{gardner2018gpytorch} based on the work presented in \citet{ryou_tal_ijrr}.

\subsection{Comparison with Prior Work}
The proposed method is built upon the MFBO framework used in~\cite{ryou_tal_ijrr, ryou_tal_corl}. 
In \citet{ryou_tal_ijrr}, MFBO is employed to create trajectories for specific waypoint sequences, while in \citet{ryou_tal_corl}, they expand this to general waypoint sequences by pretraining with a MFBO-labeled subdataset and fine-tuning through RL.
This current work introduces significant advancements, particularly in training efficiency and real-time re-planning. 
We integrate MFBO with the RL training process, effectively shortening training time and enabling real-world experiment inclusion, contrasting the previous simulation-only approach.
The model proposed in this study uses the current trajectory state as an input to the planning policy, enabling trajectory planning from non-stationary states, crucial for re-planning in mid-course.
Additionally, our planning model directly determines the absolute time for waypoint traversal, unlike the previous method's use of proportional time distributions. 
This eliminates the need for extra line-search procedures to maintain trajectory feasibility, thereby enhancing the computational efficiency of trajectory generation.

% !TeX root = ../root.tex

\section{Algorithm} \label{sections:algorithm}
\subsection{Problem Definition}

\begin{figure*}[]
\centering
\includegraphics[width=0.98\textwidth,trim=0.cm 0.cm 0.cm 0.cm,clip]{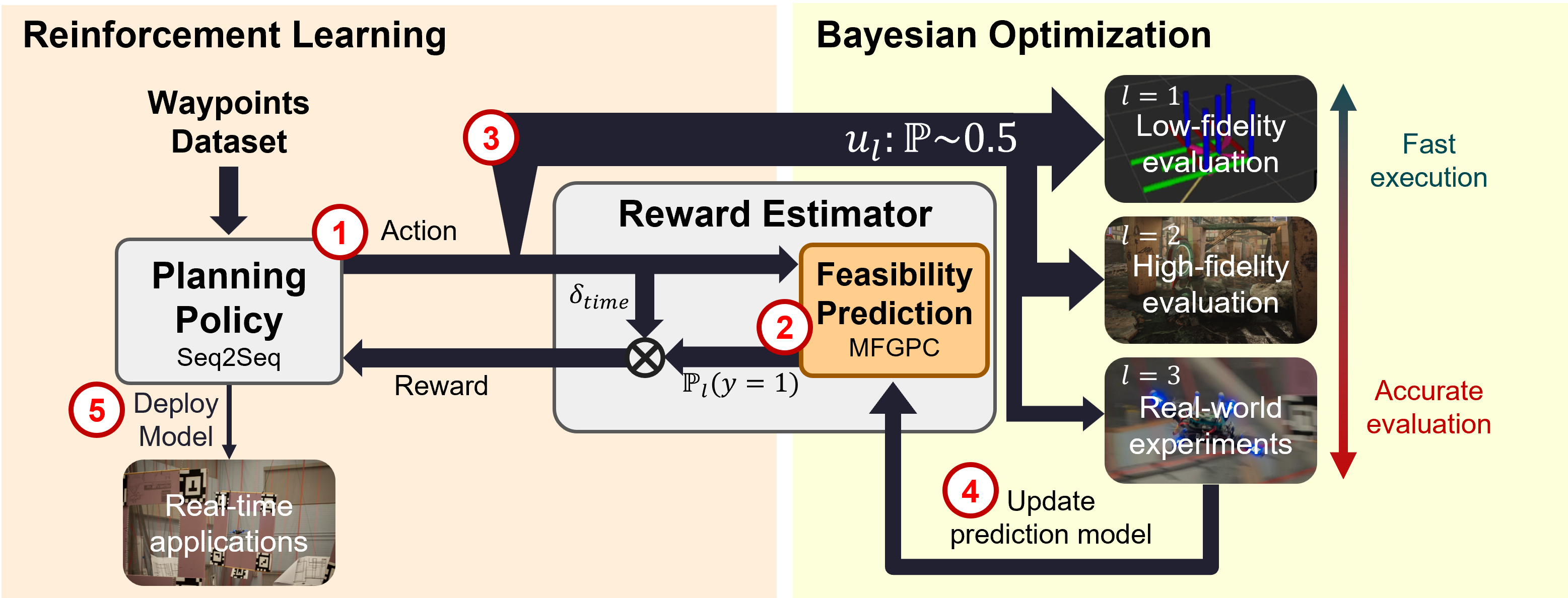}
\caption{
Overview of the proposed multi-fidelity reinforcement learning (MFRL) method:
(1) Planning policy generates actions (time allocation and smoothness weights) for waypoint sequences.
(2) Reward estimator predicts trajectory feasibility and estimates rewards, where reward is the product of feasibility prediction and time reduction achieved.
(3) After updating the planning policy, portions of the training batch with high uncertainty in feasibility prediction are selected for further multi-fidelity evaluation.
(4) Evaluation samples vary by computational cost of each fidelity level; results update the feasibility prediction model.
(5) Iterative updates train a policy maximizing time reduction while maintaining feasibility. 
The method incorporates real-world experiments for direct deployment in real-world online planning applications.
}
\label{fig:mfrl_alg}
% \vspace{-1em}
\end{figure*}

Our objective is to develop an online planning policy model capable of generating a time-optimal quadrotor trajectory that connects updated waypoints.
As shown in Figure \ref{fig:intro_main}, we consider the scenario in which the vehicle follows a trajectory $p(t) = \begin{bmatrix} {{}p_r(t)}^\intercal &  p_\psi(t) \end{bmatrix}^\intercal$ that traverses the $m+1$ prescribed waypoints $\tilde {\mathbf{p}} =  \begin{bmatrix}\tilde p^0 & \cdots& \tilde p^m\end{bmatrix}$, and the waypoints are updated as $\tilde {\mathbf{p}}_{new} = \begin{bmatrix}\tilde p^{0} & \cdots& \tilde p^k \quad \tilde p_{new}^{k+1} \cdots \tilde p^m_{new}\end{bmatrix}$.
We assume that the vehicle is located between $k-1$-th and $k$-th waypoints, and the waypoints are updated from the $k+1$-th waypoint onwards.
The time-optimal online planning problem is then formulated to minimize the traversing time of the updated trajectory $p_{new}(t)$ as follows:
\begin{equation}
\begin{aligned}
&\underset{p_{new}}{\text{minimize}} \; T_{new, m} \;\;\; \text{subject to}\\
&p_{new}(T_{new, i}) = \tilde{p}^i_{new}, \; i=k,\dots,m, \\
&\mathcal{D\,} p_{new}(T_{k}) = \mathcal{D\,} p(T_{k}), \\
&p_{new} \in \mathcal{P}(\mathcal{D\,} p(T_{k})),
\label{eqn:planning_general}
\end{aligned}
\end{equation}
where $T_i$ and $T_{new, i}$ are the reference times of the original and updated trajectory, respectively, at the $i$-th waypoint ($T_{new, i} = T_i \; \forall i=0,\cdots,k$).
$\mathcal{D\,} p(t) = \begin{bmatrix}p_r^{(1)}(t) & \cdots & p_r^{(4)}(t) & p_\psi^{(1)}(t) & p_\psi^{(2)}(t)\end{bmatrix}$ 
is the set of all derivatives at time $t$, and $\mathcal{P}(\mathcal{D\,}p)$ denotes the set of feasible trajectories that starts from the reference state $\mathcal{D\,} p$.
All initial trajectories begin in a stationary state, with all elements of $\mathcal{D\,}p(0)$ at zero, and both initial and adapted trajectories end in a stationary state, though potentially at different positions.
Furthermore, we account for scenarios where $k=0$, allowing the same planning policy to be applied for generating trajectories from the start.

The snap minimization method can be used to convert the quadrotor trajectory generation problem into a finite-dimensional optimization formulation~\cite{mellinger2011minimum, richter2016polynomial, ryou_tal_corl}.
% By minimizing the fourth-order derivative of position and the second-order derivative of yaw, it generates a smooth trajectory from a given time allocation.
In our work, we employ the extended minimum snap trajectory formulation~\cite{ryou_tal_corl}, which includes smoothness weights between polynomial pieces in the time allocation parameterization.
The smoothness weights selectively relax snap minimization objective between trajectory segments, increasing the representability of the trajectory and enabling more aggressive maneuvers.
We define $\chi(\mathbf{x}, \mathcal{D\,} p(T_{k}), \tilde {\mathbf{p}}_{new})$ as a function that maps time allocation and smoothness weights to the polynomial trajectory, which is obtained by solving the following quadratic programming problem:
\begin{equation}
\begin{gathered}
\underset{p_{new}}{\text{minimize}} \; \sum_{i=k+1}^{m} x_{wi} \int_{T_{new, i-1}}^{T_{new, i}} \mu_r \norm{p_{r}^{(4)}}^2 + \mu_\psi \Big(p_\psi^{(2)}\Big)^2 dt \\
\text{subject to} \;\;\; p_{new}(T_{new, i}) = \tilde{p}^i_{new}, \; i=k,\dots,m, \\
\mathcal{D\,} p_{new}(T_{k}) = \mathcal{D\,} p(T_{k}),
\label{eqn:alg_minsnap}
\end{gathered}
\end{equation}
where $\mu_r$ and $\mu_\psi$ are the weighing parameters.
The optimization variable $\mathbf{x} = [\mathbf{x}_t,\mathbf{x}_w]$ is composed of the time allocation $x_{t,i}$ $(T_{new, i}=T_{k} + \sum_{j=k+1}^{i} x_{t,j})$ and the smoothness weight $x_{w,i}$ $(\sum_{j=k+1}^{m} x_{w,j} = 1)$ on $i$-th polynomial piece connecting waypoints $\tilde p_{i-1}$ and $\tilde p_i$.

In this paper, we present a multi-fidelity reinforcement learning method that solves the following optimization problem:
\begin{equation}
\begin{aligned}
&\underset{\mathbf{x} \in \realR^{2 \times (m-k)}_{\geq 0}}{\text{minimize}} \; T_{new, m}, 
\; \text{subject to} \\ 
&\chi(\mathbf{x}, \mathcal{D\,} p(T_{k}), \tilde{\mathbf{p}}_{new}) \in \mathcal{P}(\mathcal{D\,} p(T_{k})).
\label{eqn:alg_minsnap_reform}
\end{aligned}
\end{equation}
Our goal is to minimize trajectory time while ensuring feasibility constraints are satisfied. 
Specifically, trajectory feasibility is defined as maintaining tracking errors within specified thresholds during real-world flight.
Unlike the snap minimization method \eqref{eqn:smoothness} that rely solely on smoothness objectives to indirectly address feasibility, the proposed method explicitly models unknown feasibility constraints $p \in \mathcal{P}(\mathcal{D}p)$ to enhance time-optimality. 
With this constraints model, we train a planning policy that generates optimal time allocations and smoothness weights for any given waypoint sequence.

The algorithm integrates two complementary components: BO and RL. 
The BO component cost-effectively creates the training dataset for modeling feasibility constraints by identifying trajectories near decision boundaries and evaluating feasibility through real-world flight tests.
To further improve efficiency, we employ MFGPC as a surrogate model, which incorporates low-fidelity evaluation results to accurately predict constraints with minimal real-world evaluations.

The RL component trains the planning policy to generate time-optimal parameters while satisfying the modeled feasibility constraints. 
It uses the expected trajectory time reduction as the reward function, calculated using the feasibility constraints model developed through BO. 
Rather than functioning as separate stages, these components function synergistically—BO leverages planning policy outputs instead of random trajectories to efficiently generate the training dataset, while RL continuously refines the policy based on the evolving constraint model.
Figure \ref{fig:mfrl_alg} illustrates an overview of the proposed multi-fidelity reinforcement learning procedure.

\subsection{Multi-Fidelity Evaluations}
\begin{figure*}[!h]
\centering
\includegraphics[width=\textwidth]{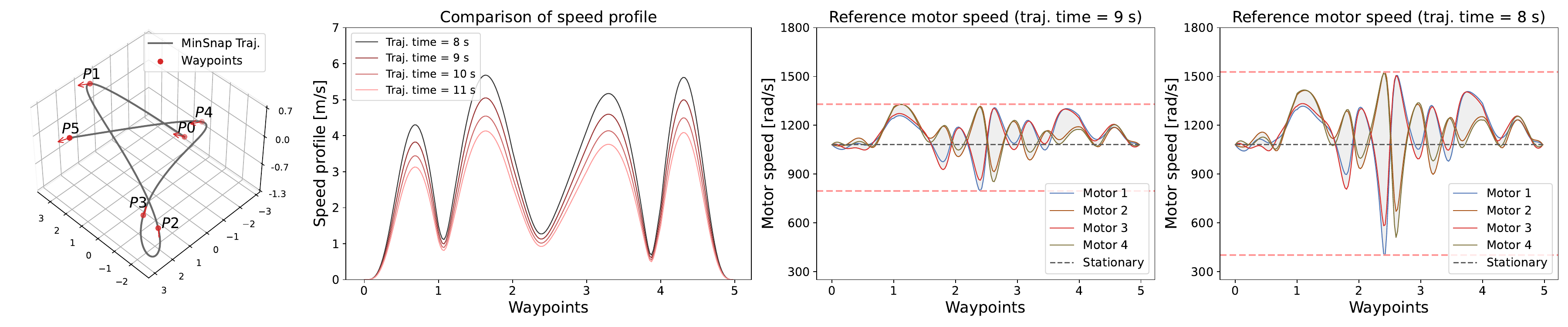}
\caption{Comparison of the speed profile and reference motor commands over trajectory time. 
The leftmost figure visualizes a minimum snap trajectory (\textit{MinSnap Traj.}), which maintains its shape with uniform scaling of time allocation. 
The second figure illustrates how reducing time allocations uniformly increases the speed profile. 
The figures on the right side compare the reference commands of the quadrotor's four motors from a minimum snap trajectory with the same waypoints but different time allocation scaling.
As time allocations reduce, the trajectory's maximum and minimum motor speeds increasingly diverge from the hovering motor speed (\textit{Stationary}).}
\label{fig:alg_eval_compare_ms}
\end{figure*}

\begin{figure}[!h]
\centering
\includegraphics[width=0.48\textwidth,trim=0.cm 0.cm 0.1cm 0.cm,clip]{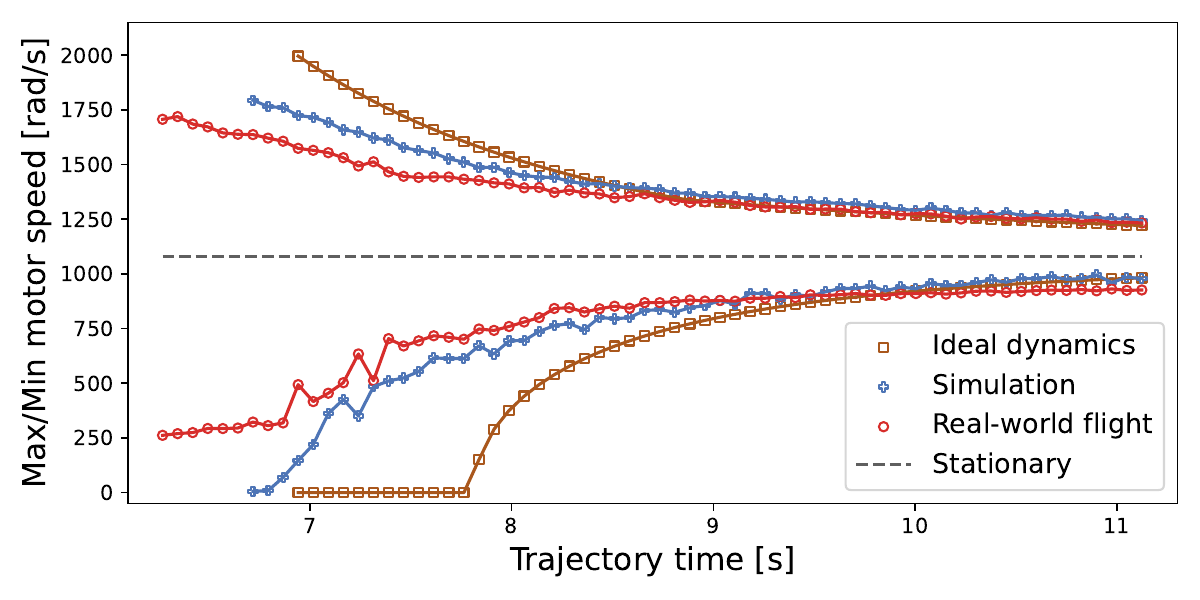}
\caption{
Variations in maximum and minimum motor speeds with time allocation scaling across different fidelity levels: \textit{Ideal dynamics}, \textit{Simulation} and \textit{Real-world flight}.
Lowering time allocations results in the trajectory's maximum and minimum motor speeds diverging from the stationary hovering motor speed (\textit{Stationary}).
At low speeds, all fidelity levels exhibit similar motor speeds. 
However, as time allocations are reduced, the fidelity gap widens, leading to noticeable differences in the feasibility boundaries across the fidelity levels.
Simulation and real-world flight cases are plotted until infeasibility or excessive tracking error occurs. 
The ideal-dynamics model is plotted further, becoming infeasible around 7.8 seconds when motor commands saturate.
}
\label{fig:alg_eval_compare}
\end{figure}
We utilize three different fidelity levels of feasibility evaluation across the paper, and our optimization goal is to satisfy the feasibility constraints at the highest fidelity level. 
The trajectory feasibility of each fidelity level is defined as inclusion in a corresponding feasible trajectory set, defined as $\mathcal{P}_{1}$, $\mathcal{P}_{2}$, and $\mathcal{P}_{3}$ respectively.
The first fidelity evaluations are based on differential flatness of the quadrotor dynamics, which enables us to transform a trajectory and its time derivatives from the output space, \ie, position and yaw angle with derivatives, to the state and control input space.
The resulting reference control input trajectory $u(t) = \zeta(p,t)$ would enable a hypothetical perfect quadrotor aircraft to track the trajectory $p$.
The set of feasible trajectories at this fidelity level is defined as
\begin{equation}
\mathcal{P}_{1} = \: \Big\{p\Big|\zeta(p,t) \in \big[\underline u, \bar u\big]^4\;\;\;
\forall t \in \left[0,T\right]\Big\},
\label{eqn:feasibility_diff_flat}
\end{equation}
where $\underline u$ is the minimum, $\bar u$ is the maximum motor speed, and $T$ is the total trajectory time.
The second fidelity evaluations are obtained using numerical simulation.
The feasible set is defined as
\begin{align}
\mathcal{P}_{2} = \Big\{p\Big|&\norm{p_r(t)-r(t)} \leq \bar r \;\land\; \nonumber \\ &|{p_\psi(t)}-{\psi(t)}| \leq \bar \psi\ \;\;\;
\forall t \in \left[0,T\right]\Big\},
\label{eqn:feasibility_tracking_err}
\end{align}
where $\bar r$ is the maximum allowable Euclidean position tracking error, and $\bar \psi$ is the maximum allowable yaw tracking error.
$r(t)$ and $\psi(t)$ are the simulated 3D position and yaw angle, respectively.
At this fidelity level, FlightGoggles~\cite{guerra2019flightgoggles}, an open-source multicopter simulator, is used with trajectory tracking control~\cite{tal2020accurate} to consider stochastic noise, motor dynamics, and aerodynamic effects.
It's worth mentioning that any vehicle dynamics and IMU simulation could be used, as our proposed algorithm is not dependent on the specific simulation or evaluation method.
Tracking error calculation requires non-parallelizable dynamics updates and controller simulations with denser time steps, significantly increasing computational time compared to the first fidelity level.
The third fidelity evaluations are obtained from real-world experiments using a quadrotor aircraft and motion capture system.
For these actual flight tests, each evaluation incorporates dynamics of the full system, including actuation and sensor systems, vehicle vibrations, unsteady aerodynamics, battery performance, and estimation and control algorithms.
We use the same controller as in simulation, and perform multiple flights to account for stochastic effects.
The feasible set $\mathcal{P}_3$ is defined identically to \eqref{eqn:feasibility_tracking_err}, but the 3D position $r_\text{mocap}(t)$ and yaw angle $\psi_\text{mocap}(t)$ are measured using the motion capture system: 
\begin{align}
\mathcal{P}_{3} = \Big\{p\Big|&\norm{p_r(t)-r_\text{mocap}(t)} \leq \bar r \;\land\; \nonumber \\ &|{p_\psi(t)}-{\psi_\text{mocap}(t)}| \leq \bar \psi\ \;\;\;
\forall t \in \left[0,T\right]\Big\},
\label{eqn:feasibility_tracking_err_3}
\end{align}
The feasibility bound can be estimated more precisely as the fidelity level increases, but the evaluation time grows exponentially.
Empirically, evaluations at the first fidelity level take about 10 milliseconds, while the second level takes roughly 2 seconds, and the third level around 2 minutes, with times changing based on trajectory lengths.
Consequently, it is impractical to directly employ the high-fidelity models in RL, which demands millions of training samples.
To address this issue, our method utilizes MFBO to efficiently construct a high-fidelity model based on the low-fidelity feasibility model.

% Simulation is slow
For slow trajectories, the differences between the three fidelity levels are minimal, but as speed increases, distinct behaviors emerge at each fidelity level.
Figure \ref{fig:alg_eval_compare} illustrates the variation in maximum and minimum motor speeds for the trajectory shown in Figure \ref{fig:alg_eval_compare_ms} across different fidelity levels.
The maximum and minimum motor speeds are critical determinants of a trajectory's feasibility because tracking errors increase when motor speed commands reach their boundaries and begin to saturate. 
As the trajectory time is reduced using the line search procedure in \eqref{eqn:minsnap_linesearch}, the maximum and minimum motor speeds increasingly diverge from the stationary motor speed, which is the speed when the vehicle maintains hovering.
At lower speeds, motor speeds are similar across all fidelity levels, but as speed increases, the motor speed at the lowest fidelity level — the reference motor speed commands — quickly diverges, whereas the motor speeds in simulations and flight experiments do so more gradually due to controllers filtering out speed command spikes.
Consequently, in the trajectory depicted in Figure \ref{fig:alg_eval_compare_ms}, the vehicle is capable of achieving higher speeds in the simulated environment compared to that estimated using the ideal dynamics model.
It is noteworthy, however, that there may be opposite scenarios where the trajectory is slower in the simulated evaluation due to the control logic.
Similarly, the discrepancies between simulated and real-world dynamics models cause the motor speeds to diverge differently between simulation and real-world scenarios. 
As trajectories's speed increases, the disparity between fidelity levels becomes more pronounced.
Optimal solutions for time-optimal trajectories are typically identified in areas where this fidelity gap is significant, highlighting the need for adaptation across different fidelity levels.

\subsection{Dataset Generation}
The training data comprises waypoint sequences commonly used in quadrotor trajectory planning, obtained by randomly sampling waypoints within topological constraints.
Initially, a dataset of waypoint sequences $\mathcal{D}_{init}$ is created by randomly sampling sequences within the unit cube $[-0.5, 0.5]^3$. 
These samples are then accepted if they satisfy two topological criteria, the total Menger curvature, \cite{menger1930untersuchungen}, and the total distance between waypoints, as follows:
\begin{align}
I_{\text{curvature}} &= {\sum \nolimits}_{i=0}^{m-2} R(\tilde{p}_r^i, \tilde{p}_r^{i+1}, \tilde{p}_r^{i+2})^{-1} \in [5, 20], \\
I_{\text{distance}} &= {\sum \nolimits}_{i=0}^{m-1} d(\tilde{p}_r^i, \tilde{p}_r^{i+1})  \in [0, 30]
\end{align}
where $R(\cdot)$ denotes the radius of the circle that passes through the three waypoints, and $d(\cdot)$ represents the Euclidean distance between two consecutive waypoints.
The sequence of sampled waypoints is temporarily connected with a minimum-snap trajectory in \eqref{eqn:minsnap_nonlinear} with no yaw objective function.
The waypoint sequences that exceed the unit cube are rejected, and the yaw component for each waypoint are assigned to be tangential to the local velocity.
The actual positions of waypoints are determined by scaling the waypoint sequence with the desired space scale $L_{\text{space}} \in \mathbb{R}^3$.
The minimum-snap time allocation ratios $\tilde{\mathbf{x}}_t^{\text{MS}}$ ($\sum_i \tilde{x}^{\text{MS}}_{t,i} = 1$) are determined by solving the nonlinear optimization \eqref{eqn:minsnap_nonlinear} with the obtained reference yaw and the stationary initial and final states.

\begin{figure}[!h]
\centering
\begin{subfigure}[b]{0.23\textwidth}
    \captionsetup{justification=centering}
    \includegraphics[width=\textwidth,trim=0.cm 0.cm 0.cm 0.cm,clip]{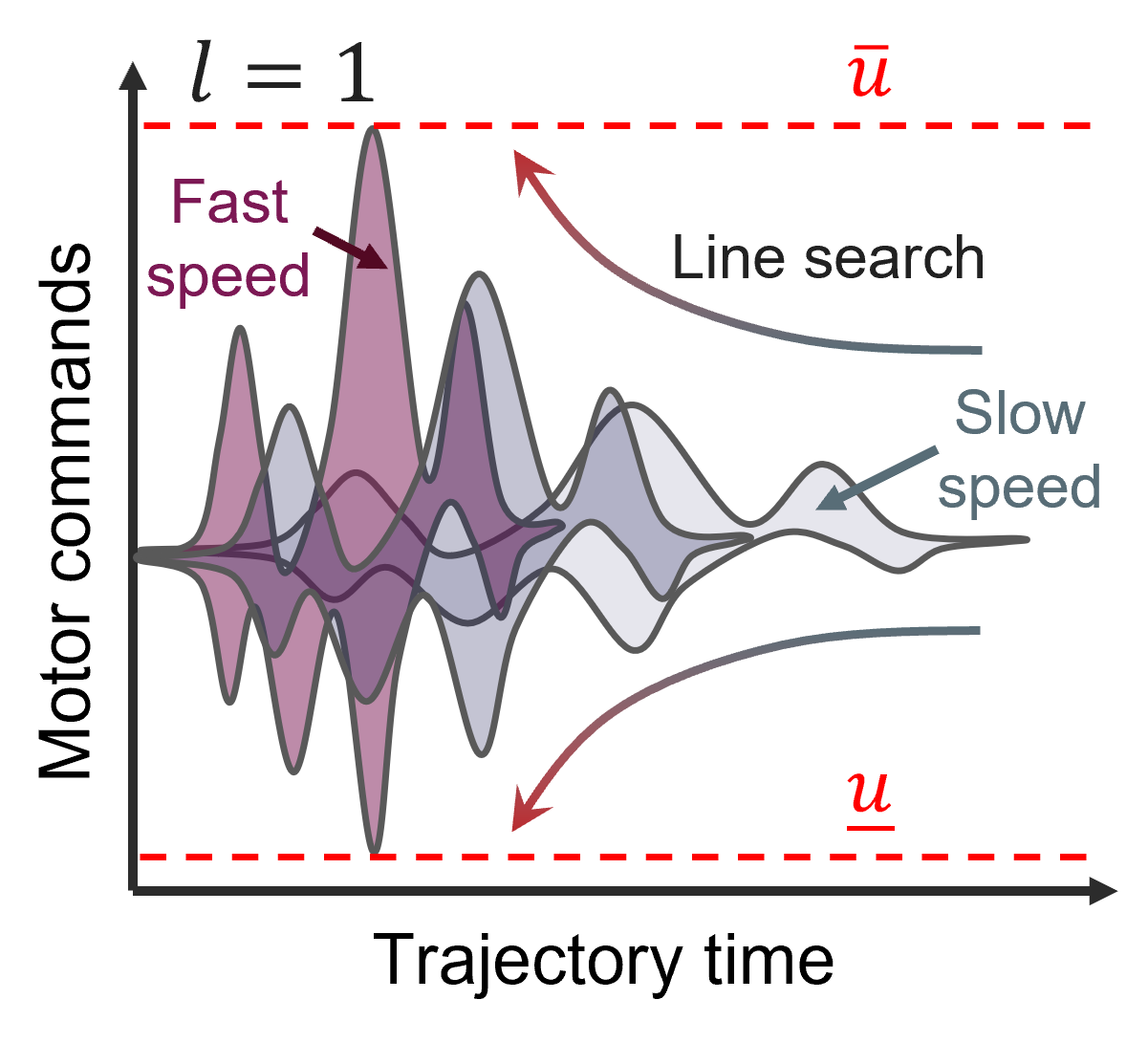}
    % \vspace{-2.em}
    % \caption{}
\end{subfigure}
\begin{subfigure}[b]{0.23\textwidth}
    \captionsetup{justification=centering}
    \centering
    \includegraphics[width=\textwidth,trim=.0cm 0.cm .0cm .0cm,clip]{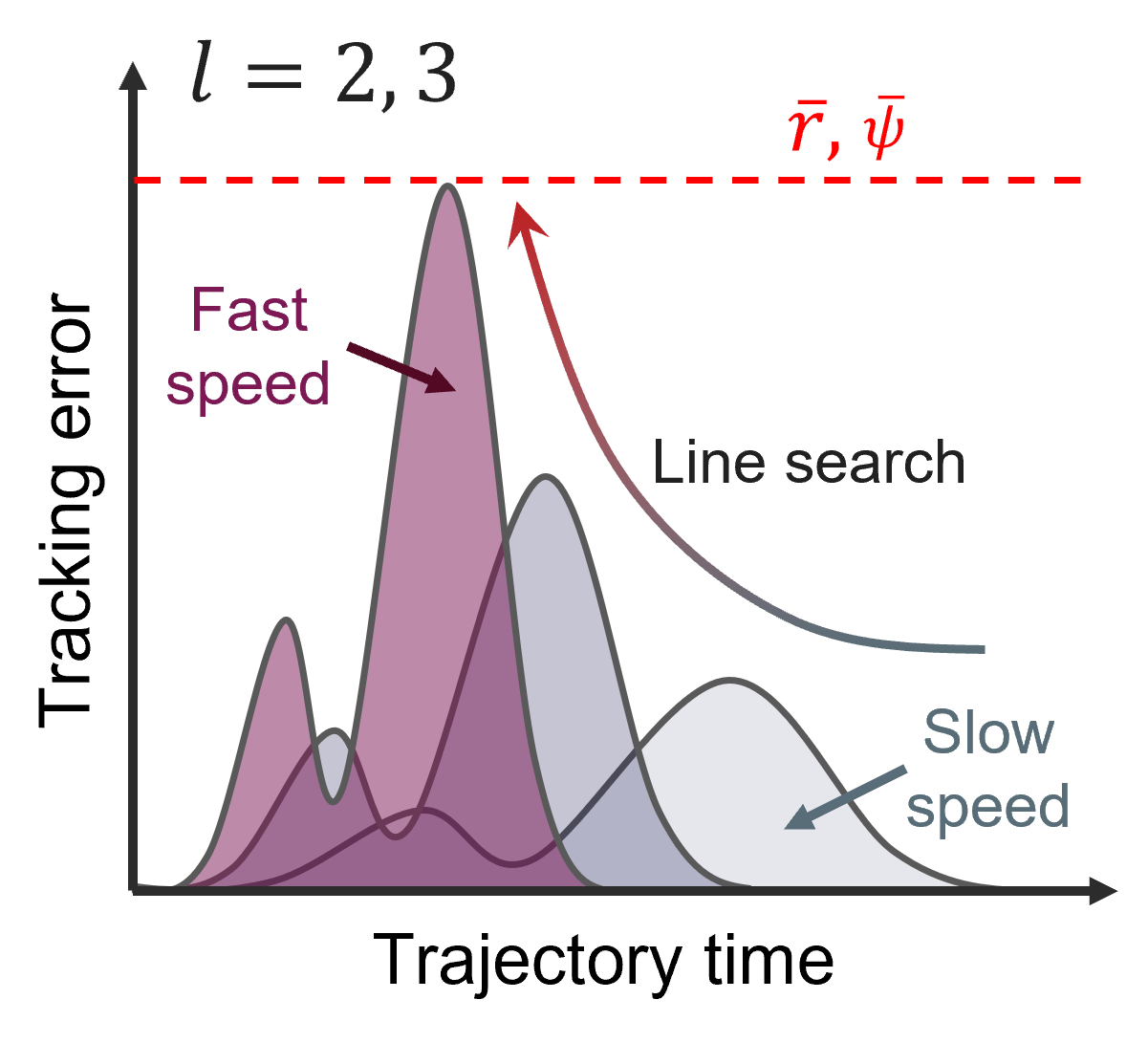}
    % \vspace{-2.em}
    % \caption{}
\end{subfigure}
\vspace{-.5em}
\caption{
Line search across fidelity levels. 
(Left) First level ($l=1$): Reduce trajectory time to motor command boundary, $[\underline u, \bar u]$. 
(Right) Second and third levels ($l=2,3$): Reduce time to tracking error threshold, $\bar r, \bar \psi$.
}
\label{fig:alg_data_gen_linesearch}
\end{figure}

\begin{figure}[!h]
\centering
\includegraphics[width=0.48\textwidth,trim=0.cm 0.cm 0.cm 0.cm,clip]{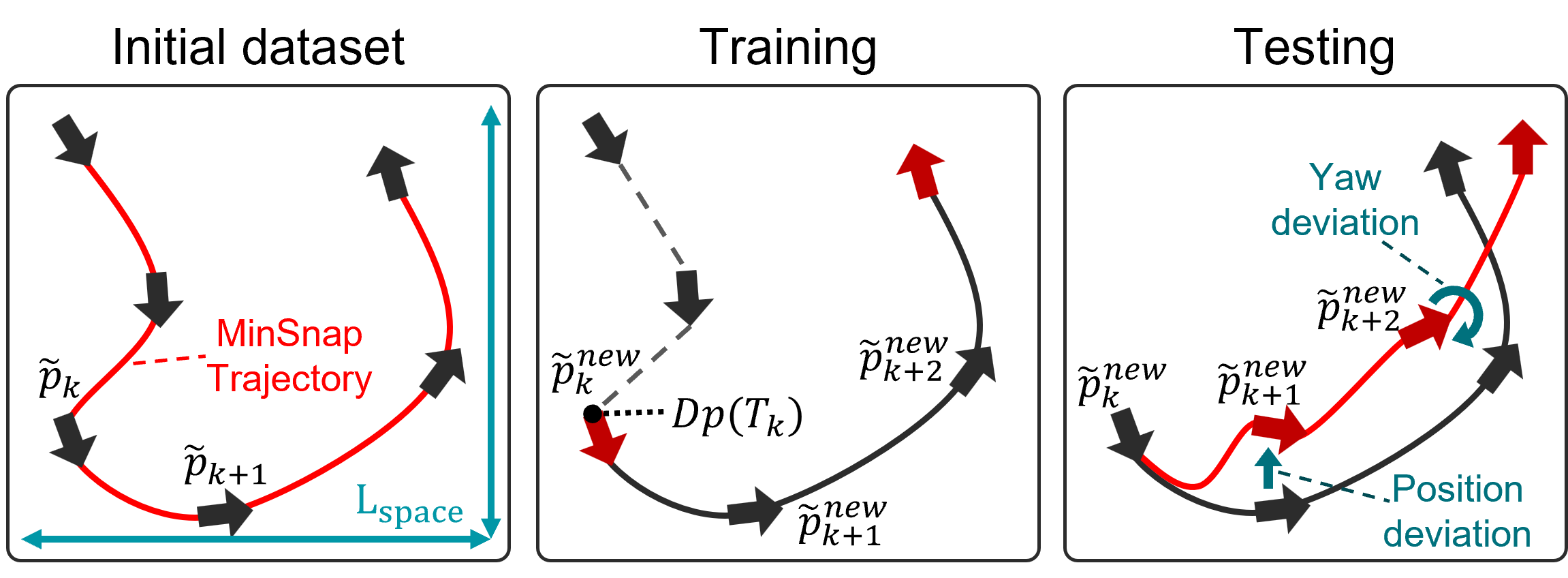}
\vspace{-.5em}
\caption{
Waypoints dataset generation and usage in training and testing. 
\textit{Initial dataset}: Randomly generated waypoints scaled to room size $L_\text{space}$, connected using snap minimization. 
\textit{Training}: Planning policy reconstructs trajectories from random midpoints of waypoint sequences; reward estimator predicts trajectory feasibility. 
\textit{Testing}: Position and yaw deviations added to waypoint sequences to evaluate policy robustness.
}
\label{fig:alg_data_gen}
\end{figure}

% Define the adapted MS
Once the time allocation ratios are determined, a line-search, as defined in \eqref{eqn:minsnap_linesearch}, is performed to find the minimal trajectory time $T^{MS}_{l}$ for each fidelity level $l\in\{1,2,3\}$ that satisfies the feasibility constraints.
Figure \ref{fig:alg_data_gen_linesearch} illustrates the line search procedure in each fidelity level.
Due to the computational expense of calculating minimal times for all waypoint sequences in the second and third fidelity levels, a subset of sequences is used to estimate the inter-fidelity ratio, which then approximates minimal times for the higher levels.
To be specific, for the lowest-fidelity level of dataset, the trajectory times are determined for the entire dataset $\mathcal{D}_{init}$, and for the rest of the fidelity level, the trajectory times for the subset of the preceding fidelity dataset are determined~($\mathcal{D}_{3} \subset \mathcal{D}_{2} \subset \mathcal{D}_{1} = \mathcal{D}_{init})$.
By comparing the trajectory times in the each fidelity dataset to the same waypoint sequence in the lowest fidelity dataset, we estimate the trajectory time ratio:
\begin{equation}
\alpha_{l/1} = \frac{1}{N_l} {\sum \nolimits}_{i \in \mathcal{D}_l} T^{\text{MS}}_{l,i} / T^{\text{MS}}_{1,i}
\end{equation}
where $N_l$ is the size of $l$-th fidelity subset $\mathcal{D}_l$.
For the lowest-fidelity level, the time between the waypoints are obtained by multiplying the trajectory time with the time allocation ratio: $\mathbf{x}^{\text{MS}}_1 = T^{\text{MS}}_1 \tilde{\mathbf{x}}^{\text{MS}}$, and for the rest of the fidelity level, the time between the waypoints are approximated as $\mathbf{x}^{\text{MS}}_l = \alpha_{l/1} T^{\text{MS}}_1 \tilde{\mathbf{x}}^{\text{MS}}$.
In the proposed simulation and real-world flight setup, $\alpha_{l/1} \in [0.95, 1.05]$ across all fidelity levels and datasets. 
This approximates time allocation for higher-fidelity levels, avoiding costly line searches over the entire dataset. 
The parameter is used only for training; for testing, we conduct line searches to determine exact minimum time allocations.
The planning policy is trained to reconstruct the trajectory starting from a randomly selected midpoint of a training waypoint sequence, while the reward estimator is trained to predict the feasibility of the reconstructed trajectory.
During the testing phase, we use the time allocation $\mathbf{x}^{\text{MS}}_l$ from the highest fidelity level solution of \eqref{eqn:minsnap_nonlinear} and \eqref{eqn:minsnap_linesearch} as the baseline. 
We also add position and yaw deviations to waypoints to assess policy robustness.
Figure \ref{fig:alg_data_gen} shows the dataset usage in training and testing.

\begin{figure*}[]
\centering
\includegraphics[width=0.96\textwidth,trim=0.cm 0.cm 0.cm 0.cm,clip]{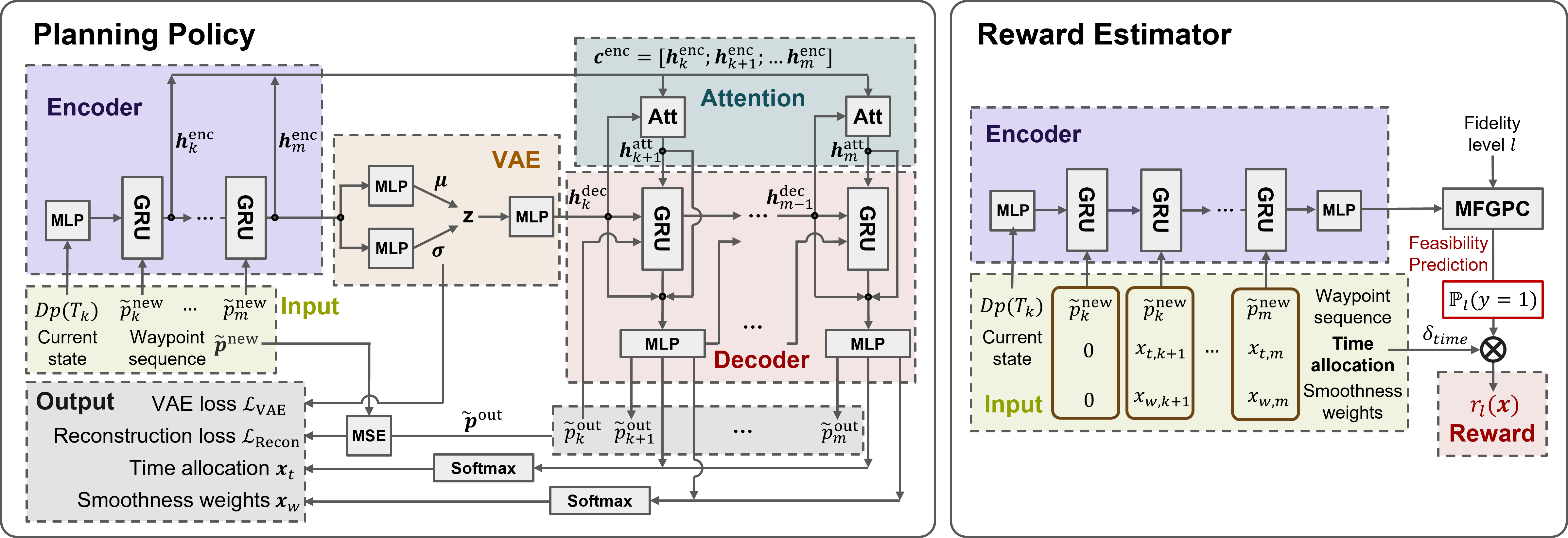}
\caption{
Overview of the planning policy and reward estimator architecture.
(Left) The planning policy, based on a sequence-to-sequence model, determines time allocation and smoothness weights for input waypoints.
The model comprises an encoder (bi-directional GRU), decoder (GRU), VAE, and attention modules. 
(Right) The reward estimator predicts feasibility of the policy outputs using a multi-fidelity Gaussian process classifier.
The reward signal is calculated as a product of feasibility prediction and time reduction achieved by the policy output.
}
\label{fig:cond_s2s}
% \vspace{-1em}
\end{figure*}

% !TeX root = ../root.tex

\subsection{Training Reward Estimator}
The reward estimator calculates the expected time reduction by predicting the feasibility of trajectories generated from given time allocations and smoothness weights $\mathbf{x} = [\mathbf{x}_t,\mathbf{x}_w]$, as follows:
\begin{equation}
    P_l(y|\mathbf{x}, \mathcal{D} p, \tilde{\mathbf{p}}) = 
    \begin{cases}
        P(\chi(\mathbf{x}, \mathcal{D\,} p, \tilde{\mathbf{p}}) \in \mathcal{P}_l(\mathcal{D\,} p)) & \text{if $y = 1$}, \\
        P(\chi(\mathbf{x}, \mathcal{D\,} p, \tilde{\mathbf{p}}) \notin \mathcal{P}_l(\mathcal{D\,} p)) & \text{if $y = 0$},
    \end{cases}
\end{equation}
where $\chi(\mathbf{x}, \mathcal{D\,} p, \tilde{\mathbf{p}})$ refers to the polynomial trajectory generated with \eqref{eqn:alg_minsnap} and $\mathcal{P}_l$ is the set of feasible trajectories at the $l$-th fidelity level.

To efficiently capture the correlation between different fidelity levels from a sparse dataset, the feasibility probability is modeled using a Gated Recurrent Unit (GRU) and a Multi-Fidelity Gaussian Process Classifier (MFGPC), as illustrated in Figure \ref{fig:cond_s2s}.
The GRU is used to extract a feature vector from the given waypoints sequence $\tilde{\mathbf{p}}$, time allocation $\mathbf{x}_t$, and smoothness weights $\mathbf{x}_w$.
The initial hidden state of GRU is obtained from the trajectory state $\mathcal{D\,} p$, and the last hidden output of GRU is converted into the feature vector through a multi-layer perceptron (MLP).
Subsequently, MFGPC serves as the final stage in the feasibility estimation pipeline, utilizing this feature vector to predict feasibility. 
The parameters of GRU and MFGPC are trained through the minimization of a variational lower bound, as detailed in \eqref{eqn:alg_modeling_mfdgp_loss_all}.

\begin{figure}[!h]
\centering
\begin{subfigure}[b]{0.23\textwidth}
    \captionsetup{justification=centering}
    \includegraphics[width=\textwidth,trim=0.cm 0.cm 0.cm 0.cm,clip]{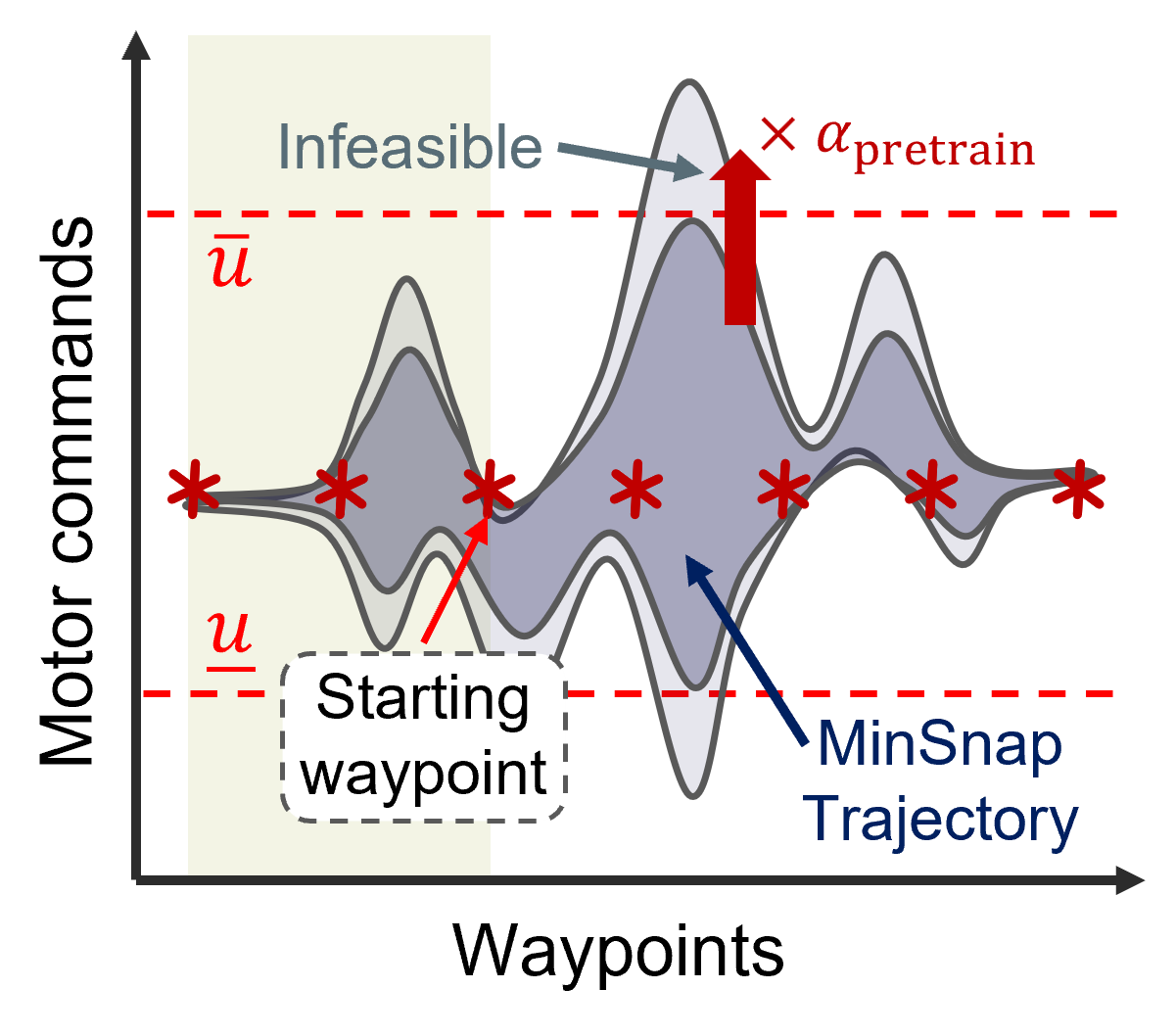}
    % \vspace{-2.em}
    % \caption{}
\end{subfigure}
\begin{subfigure}[b]{0.23\textwidth}
    \captionsetup{justification=centering}
    \centering
    \includegraphics[width=\textwidth,trim=.0cm 0.cm .0cm .0cm,clip]{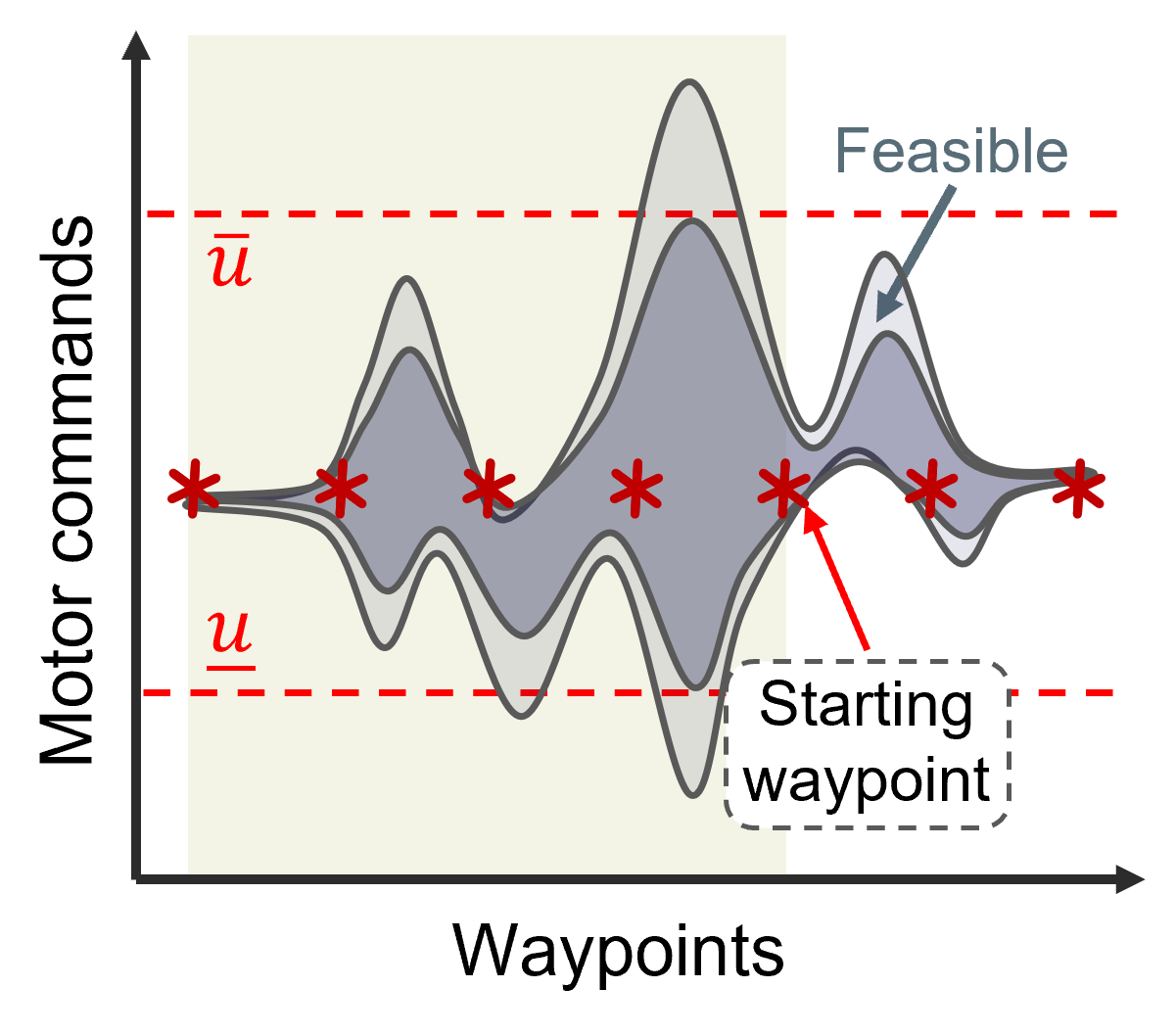}
    % \vspace{-2.em}
    % \caption{}
\end{subfigure}
\vspace{-.5em}
\caption{
Trajectory labeling for reward estimator pretraining. 
(Left) Trajectories from snap minimization (dark blue region) are sped up to their feasibility limit; further acceleration makes them infeasible (grey region). 
The yellow shaded region represents the already traversed portion of the trajectory.
(Right) After scaling, most trajectories become infeasible, but some remain feasible when reconstructed from random midpoints in the training data.
}
\label{fig:alg_pretrain_label}
\end{figure}

% Pretraining
The feasibility prediction model is pretrained using a dataset with fictitious labels.
We randomly select half of the $l$-th fidelity dataset $\mathcal{D}_{l}$ and reduce its trajectory time $T^{\text{MS}}_l$ by $\alpha_{\text{pretrain}} = 0.9$.
Although the data points with the reduced trajectory time may be feasible, as illustrated in Figure \ref{fig:alg_pretrain_label}, they are currently labeled as infeasible, while the rest retain their feasible labels.
This approach simulates infeasible data without risking actual infeasible trajectory generation, as all original trajectories are from the feasible set near the boundary. 
The model learns to approximate the feasibility boundary based on the principle that slower trajectories are generally more feasible. 
Once the model learns a brief structure of the feasibility boundary, it can be refined during the following MFRL procedure, which safely and efficiently collects a new dataset from scratch near the boundary.
The time scaling factor $\alpha_{\text{pretrain}}$ balances two extremes: if too small, the model can't differentiate near-boundary trajectories; if too large, it lacks data at speed extremes. 
This parameter is empirically chosen to maximize MFRL training stability, as extreme values may cause the model to converge to suboptimal solutions.
For training, waypoint sequences are inputted to the model starting from randomly selected midpoints as shown in Figure \ref{fig:alg_data_gen}.

During the RL training, we employ BO to train the reward estimator. 
Specifically, the reward estimator chooses a subset of the dataset along with the corresponding policy outputs and annotates them with multi-fidelity evaluations.
To be specific, policy outputs with a high level of uncertainty are selected, where the uncertainty is quantified using variation ratios~\cite{freeman1965elementary}:
\begin{align}
u_l(\mathbf{x}_t, \mathbf{x}_w) = 1 - {\max \nolimits}_{c\in\{0,1\}} P_l(y=c|\mathbf{x}, \mathcal{D} p, \tilde{\mathbf{p}}).
\label{eqn:alg_acquisition_explore}
\end{align}
The number of selected samples varies between the fidelity levels, with more samples chosen for low-fidelity evaluations and only a few for high-fidelity evaluation due to the associated cost. 
MFGPC compensates for the imbalanced sample size by modeling the correlation between the different fidelity levels.
After evaluations are completed, the feasibility model is updated using a training dataset that includes the newly assessed data points and randomly selected prior evaluations.
The subset selection and reward estimator training are performed sequentially across each fidelity level; the same data points may be evaluated at multiple fidelity levels.
At the first iteration of the RL training, to enhance training stability, policy outputs closest to the minimum-snap time allocation are chosen instead of those with high uncertainty.
% To improve the training process's efficiency, the selected data points are routed to a separate process running on the external CPU server and evaluated in parallel with the policy update procedure.

Finally, the expected time reduction is estimated by combining the relative time reduction with the feasibility probabiliy:
\begin{equation}
    r_l(\mathbf{x}) = \mathbb{E}_{P_l(y=1|\mathbf{x}, \mathcal{D} p, \tilde{\mathbf{p}})} [\bar \delta_{time, l}(\mathbf{x})]
\end{equation}
where the relative time reduction $\bar \delta_{time, l}$ is obtained by comparing the time allocation $\mathbf{x}_{t}$ with minimium snap time allocation: 
\begin{align}
\delta_{time, l}(\mathbf{x}) &= 1 - \left({\sum \nolimits}_{i=k+1}^m x_{t,i}\right) / \left(\alpha_{l/1} T^{\text{MS}}_1 {\sum \nolimits}_{i=k+1}^m \tilde{x}^{\text{MS}}_{t,i}\right) \nonumber \\
\bar \delta_{time, l}(\mathbf{x}) &= \max(\delta_{time, l}(\mathbf{x}), r_{max}) + r_{bias}.
\end{align}
To prevent the policy from converging to local minima, which continue to generate infeasible solutions with extremely short trajectory times, the time reduction is transformed with the parameters $r_{max}, r_{bias} \in [0,1]$.
% $r_{max}$ and $r_{bias}$

% !TeX root = ../root.tex

\subsection{Training Planning Policy}
% The planning policy is trained to maximize the expected time reduction which is obtained from the reward estimator model.
% % In every iteration, the policy is updated using a reward estimator model, tailored to accommodate different fidelity levels. 
% The training dataset is generated by randomly slicing the waypoint sequences of the lowest-fidelity dataset $\mathcal{D}_{init}$ after scaling its time allocation with the trajectory time ratio $\alpha_{l/1}$.

The planning policy determines time allocations and smoothness weights from a sequence of remaining waypoints and a current trajectory state. 
Figure \ref{fig:cond_s2s} illustrates the planning policy, a sequence-to-sequence model comprising an encoder (bi-directional GRU), decoder (GRU), VAE, and attention modules. 
To handle variable sequence lengths, we use the sequence-to-sequence language model from \cite{cho2014proceedings} as the main module. 
To prevent memorization, a variational autoencoder (VAE) connects the encoder and decoder hidden states, densifying the hidden state into a low-dimensional feature vector as described in \cite{bowman2015generating}. 
We employ content-based attention~\cite{bahdanau2015neural} to capture the global shape of waypoint sequences, calculating similarity between encoder and decoder hidden states, $\mathbf{h}^\text{enc}_{j}$ and $\mathbf{h}^\text{dec}_{i}$, as follows:
\begin{align}
	&\mathbf{h}^{\text{att}}_{i, j} = \text{softmax}_j(\mathbf{v}_\text{att} \tanh(\mathbf{W}_\text{att} [\mathbf{h}^\text{enc}_{j}; \mathbf{h}^\text{dec}_{i}])), \nonumber \\
	&\mathbf{h}^{\text{att}}_{i} = \textstyle \sum_j \mathbf{h}^{\text{att}}_{i, j} \mathbf{h}^{\text{enc}}_{j},
\end{align}
where $\mathbf{v}_\text{att}$ and $\mathbf{W}_\text{att}$ are the weight matrix of the attention module. 
The non-linear activation function is chosen to optimize training stability.
The waypoint sequence is transformed into a fixed-dimensional feature vector through the encoder, and the initial hidden state of the encoder GRU is derived from the current trajectory state $\mathcal{D\,} p$.
The decoder output is then transformed into a latent time allocation $\tilde{\mathbf{x}}_t$ and a smoothness weight $\tilde{\mathbf{x}}_w$ via fully-connected layers (MLP).
We manage separate policy models, $\pi_l$, for different fidelity levels, but each model shares most parameters except for the last MLP layer.
To elaborate, the full architecture features one shared sequence-to-sequence backbone and three MLP heads, and this entire system—backbone and all heads—is jointly trained with the reward estimator.
The latent smoothness weight undergoes normalization via a softmax function, while the latent time allocation is rescaled using the average time after an exponential function is applied: 
\begin{align}
&\mathbf{x}_{t} = T_{avg}\exp(\tilde{\mathbf{x}}_{t})/\text{dim}(\tilde{\mathbf{x}}_{t}), \label{eqn:alg_policy_sofmax_time}\\
&\mathbf{x}_{w} = \text{softmax}(\tilde{\mathbf{x}}_{w}),
\end{align}
where $\text{dim}(\tilde{\mathbf{x}}_{t})$ refers to the number of the remaining waypoints.
The average time $T_{avg}$ is obtained by dividing the distance between remaining waypoints by the average speed of training sequences $v_{avg}$, $T_{avg} = \sum_{i=k}^{m-1} \norm{\tilde{p}^{i+1}_{r}-\tilde{p}^{i}_{r}} / v_{avg}$.
To improve training stability, the decoder also outputs reconstructed waypoints and is trained to minimize waypoint reconstruction loss: $\mathcal{L}_{\text{Recon}} = \lVert \tilde{\mathbf{p}}^\text{new} - \tilde{\mathbf{p}}^\text{out}\rVert^2$, which is the mean square error between the input and output waypoint sequences.
The outputs of the planning policy are used to generate a polynomial trajectory, using the quadratic programming in \eqref{eqn:alg_minsnap}.

Before the RL training, the planning policy is pretrained to reconstruct minimum-snap time allocations starting from a randomly selected midpoint in the sequence by minimizing
\begin{align}
	&\mathcal{L}_{\text{Pretrain}} = \mathcal{L}_{\text{Recon}} + \mathcal{L}_{\text{VAE}} \nonumber \\ 
    &\; + {\sum \nolimits}_{l=1}^{L} \left(\lVert\mathbf{x}_{l,t} - \alpha_{l/1} T^{\text{MS}}_1 \tilde{\mathbf{x}}_t^{\text{MS}}\rVert^2 + \lVert\mathbf{x}_{l,w} - \mathbf{x}_w^{\text{MS}}\rVert^2\right),
	\label{eqn:pretraining_loss}
\end{align}
where $\mathcal{L}_{\text{VAE}}$ is the evidence lower bound (ELBO) loss from the VAE~\cite{kingma2013auto}, $\mathcal{L}_{\text{Recon}}$ is the waypoint reconstruction loss, and $\mathbf{x}_t, \mathbf{x}_w \in \mathbb{R}^{m-k}_{>0}$ represent the planning policy outputs for the input sequence started from the $k$-th waypoints.
The smoothness weights are trained to output the normalized inverse of the time allocations, $x_{w,i}^{\text{MS}} = ({\textstyle\sum\nolimits}_{j=k+1}^{m} \nicefrac{\tilde{x}_{t,i}^{\text{MS}}}{\tilde{x}_{t,j}^{\text{MS}}})^{-1}$, to prevent the fully connected layer associated with the smoothness weights from becoming zeros.

\begin{figure}[]
\centering
\includegraphics[width=0.48\textwidth]{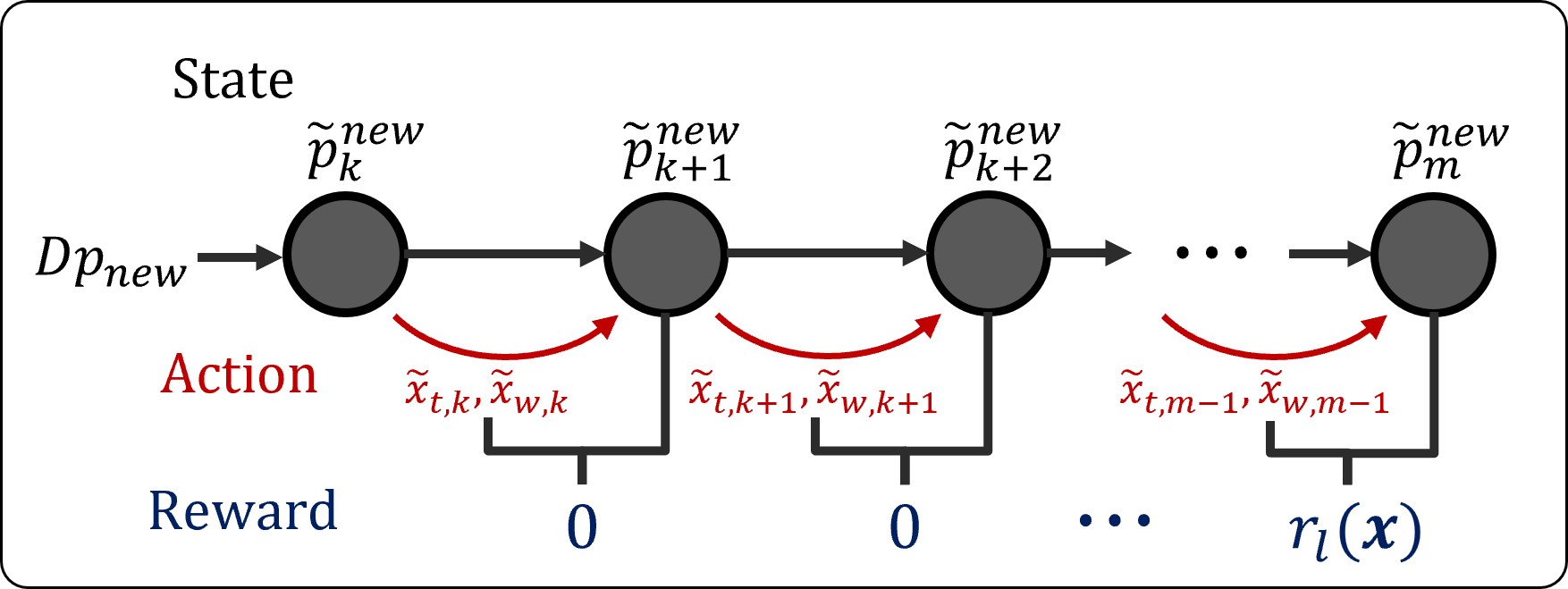}
\caption{
Structure of RL episode. 
The policy outputs time allocation and smoothness weight between consecutive waypoints. 
The agent receives a reward at episode end, based on estimated time reduction from all actions. 
The $l$-th fidelity policy uses the corresponding fidelity reward estimator.
}
\label{fig:alg_rl_episode}
\end{figure}

% Based on the low-fidelity time allocation of the lowest-fideltiy dataset $\mathcal{D}_{init}$ and the trajectory time ratio $\alpha_{l/1}$,
The planning policy is trained to maximize the expected time reduction which is obtained from the reward estimator model.
As illustrated in Figure, \ref{fig:alg_rl_episode}, we formulate a Markov decision process using the remaining waypoint as a state variable, $s_i$ and the unnormalized output of planning policy $\tilde{x}_{t,i}$ and $\tilde{x}_{w,i}$ as an action $a_i$:
\begin{equation}
s_i \coloneq \tilde{p}^\text{new}_i,\;\; a_i \coloneq [\tilde{x}_{t,i}, \tilde{x}_{w,i}]
\end{equation}
At the end of each episode, the actual time allocation is calculated using \eqref{eqn:alg_policy_sofmax_time}, which is used to estimate the expected time reduction $r_l(\mathbf{x})$ from the reward estimator.
$r_l(\mathbf{x})$ is estimated using the corresponding level's feasibility prediction and time reduction based on its minimum snap time allocation.
Since the reward is zero except at the end of the episode, the discounted reward is used for the $i$-th step reward: $r_{l,i} = \gamma^{m-i} r_l(\mathbf{x})$ ($i=k+1,\dots,m$), where $\gamma \in [0,1]$ is the discount factor.
At the end of each training epoch, rewards are batch-wise normalized by subtracting their mean and dividing them by the standard deviation.
It is noteworthy that all time allocations and smoothness weights can be considered as a single action, formulating the optimization as an RL problem with a single rollout. 
Instead, we frame it as a multiple rollout RL problem, which improves training stability although this formulation doesn't strictly satisfy the Markov condition.

The proximal policy optimization (PPO) algorithm~\cite{schulman2017proximal} is used to update the planning policy.
At each training iteration, the action 
% $\tilde{a}_i \sim \mathcal{N}(a_i, \sigma_{rf})$ 
$\tilde{a}_i \sim \pi_l (\cdot | a_i)$ 
is sampled from the normal distribution around the policy output $a_i$.
We employ PPO without a value network since training a separate value network cannot adequately account for the reward bias when estimating rewards from the continually updated feasibility boundary model.
% Training a separate value network cannot successfully capture the reward bias because the reward is estimated from the feasibility boundary model, which is kept updated throughout training.
% To estimate the advantage, we instead use the baseline reward $r_b$, which is obtained by averaging the estimated rewards at the end of each epoch: $r_b = {N_{\text{batch}}}^{-1} \sum_{n=1}^{N_{\text{batch}}} \sum_{i=k+1}^{m} r_i$ where $N_{\text{batch}}$ is the size of batch.
The reward is estimated based on the action from the old policy $\pi^\text{old}_{l}$, and the next planning policy $\pi_l$ is updated in the sense that maximizes the PPO clipping reward:  
% \begin{align}
% &r^\text{clip}_{i} = \min \left(r_{\pi,i}r_i, \text{clip}(r_{\pi,i}, 1-\epsilon, 1+\epsilon) r_i\right), \nonumber \\
% &r^\text{entropy}_{i} = -\log \pi (\tilde{a}_i | a_i), \nonumber \\
% &\mathcal{L}_{\text{Policy}} = \sum\nolimits_{l=1}^{L} \sum\nolimits_{i=k+1}^{m} r^\text{clip}_{l,i} + r^\text{entropy}_{l,i},
% \label{eq:rl_objective}
% \end{align}
\begin{align}
&\mathcal{L}^\text{Clip}_{l,i} = \min \left(\omega_{l,i}\:r_{l,i},\;\text{clip}(\omega_{l,i}, 1-\epsilon, 1+\epsilon) r_{l,i}\right), \nonumber \\
&\mathcal{L}^\text{Entropy}_{l,i} = -\log \pi_l (\tilde{a}_i | a_i), \nonumber \\
&\mathcal{L}_{\text{Policy}} = \sum\nolimits_{l=1}^{L} \sum\nolimits_{i=k+1}^{m} \mathcal{L}^\text{Clip}_{l,i} + \mathcal{L}^\text{Entropy}_{l,i},
\label{eq:rl_objective}
\end{align}
where $\omega_{l,i} = \pi_l (\tilde{a}_i | a_i) / \pi^\text{old}_{l} (\tilde{a}_i | a_i)$ is the probability ratio between the new policy and old policy.
% $$
% \mathcal{L}_{\text{Clip}} = {N_{\text{batch}}}^{-1} \sum\nolimits_{k=1}^{N_{\text{batch}}} \sum\nolimits_{i=1}^{m-k} 
% \min \left(r_\pi (r_i - r_b), \text{clip}(r_\pi, 1-\epsilon, 1+\epsilon)(r_i - r_b)\right),
% $$
The PPO objective function is minimized together with the reconstruction loss and the VAE variational loss to stablize the training procedure \ie,
\begin{equation}
	\mathcal{L}_{\text{RL}} = \mathcal{L}_{\text{Recon}} + \mathcal{L}_{\text{VAE}} + \mathcal{L}_{\text{Policy}}
\end{equation}
% where $\mathcal{L}_{\text{Ent}}$ is the entropy bonus that is estimated from the action variance $\sigma_{rf}$.

\subsection{Balancing Time Reduction and Tracking Error}
\label{subsec:alg_scaling}
In addition to the training procedure of planning policy, we propose a method to effectively balance the time reduction and the tracking error of the output trajectory. 
The trained trajectory aims to operate within the tracking error bounds established during training, but it may be necessary to readjust the trajectory's flight time and tracking error balance according to the target environment. 
For example, in obstacle-rich environments, precise tracking is critical and the trajectory's speed must be reduced, whereas in opposite scenarios, the trajectory's speed can be increased while accepting a certain degree of tracking error.

We leverage the property that the trajectory's shape obtained from the quadratic programming in \eqref{eqn:alg_minsnap} remains invariant when both time allocation and the current trajectory state are scaled.
To be specific, the polynomial $\chi(\alpha \mathbf{x}, \alpha^{-1}\mathcal{D\,} p, \tilde {\mathbf{p}})$ maintains a consistent shape across all scale factors $\alpha\in\mathbb{R}_{>0}$ where $\alpha^{-1}\mathcal{D\,} p$ is defined as
\begin{equation}
% \alpha \mathcal{D\,} p(t) \coloneqq [\alpha p_r^{(1)}(t), \dots, \alpha^4 p_r^{(4)}(t), \alpha p_\psi^{(1)}(t), \alpha^2 p_\psi^{(2)}(t)]
% \alpha^{-1} \mathcal{D\,} p \coloneqq [\alpha^{-1} p_r^{(1)}(t), \dots, \alpha^{-4} p_r^{(4)}(t), \alpha p_\psi^{(1)}(t), \alpha^2 p_\psi^{(2)}(t)]
\alpha^{-1} \mathcal{D\,} p \coloneqq \begin{bmatrix}\alpha^{-1} p_r^{(1)} \kern2pt \cdots \kern2pt \alpha^{-4} p_r^{(4)} \kern2pt \alpha^{-1} p_\psi^{(1)} \kern2pt \alpha^{-2} p_\psi^{(2)} \end{bmatrix}.
\end{equation}
For $\alpha > 1$, this transformation slows down the trajectory without altering its shape, enabling the reference motor commands to shift towards the hovering motor speed.
Building upon this transformation, we initially scale the trajectory state by multiplying the scale factor as $\mathcal{D\,}p \rightarrow \alpha \mathcal{D\,}p$ and utilize the trained policy $\pi$ to generate feasible time allocations for the scaled trajectory state: $\mathbf{x}_t, \mathbf{x}_w = \pi (\tilde{\mathbf{p}}, \alpha \mathcal{D\,}p)$.
The final trajectory is then generated with the policy output $[\alpha\mathbf{x}_t,\mathbf{x}_w]$ from the current trajectory state $\mathcal{D\,}p$.
It is noteworthy that the trajectory generated with the time allocation $\alpha \mathbf{x}_t$ and the trajectory state $\mathcal{D\,}p$ preserves the same shape but the speed profile is scaled with $\alpha^{-1}$ compared to the trajectory with $\mathbf{x}_t$ and $\alpha \mathcal{D\,}p$.
Without needing additional planning policy inference, this transformation effectively balances time reduction and tracking error by scaling the speed profile while preserving the trajectory's shape.

% !TeX root = ../root.tex

\section{Experimental Results} \label{sections:experiment}

The proposed algorithm is evaluated by training the planning policy and testing the output of the planning policy model with flight experiments.
Three different fidelity levels of evaluations are used to create the training dataset for the reward estimator.
For the first fidelity level, the reference motor commands are obtained using the idealized quadrotor dynamics-based differential flatness transform presented in \cite{mellinger2011minimum}.
The feasible set $\mathcal{P}_1(\mathcal{D\,} p)$ is defined as the set of all trajectories that start from the state $\mathcal{D\,} p$ and have reference motor commands that fall within the admissible range, $[\underline u, \bar u] = [0, 2200]\:\si{rad/s}$.
In the second and third fidelity level, the feasible sets $\mathcal{P}_2(\mathcal{D\,} p)$ and $\mathcal{P}_3(\mathcal{D\,} p)$ include all trajectories that can be flown using the flight controller~\cite{tal2020accurate} with position tracking errors less than the maximum allowable value of $\bar r = 20\:\si{cm}$, and yaw tracking errors less than $\bar \psi = 15\:\si{deg}$.
The flight experiment at the second fidelity level is performed using a 6DOF flight dynamics and inertial measurement simulation~\cite{guerra2019flightgoggles}, whereas the flight experiment at the third fidelity level is conducted using the actual vehicle.
Since the planning policy generates a trajectory starting from a midpoint in a waypoint sequence, the tracking error is estimated by simulating or flying the original trajectory to the midpoint and then switching to the updated trajectory.
Initially, we conduct a comprehensive analysis of the proposed algorithm in a simulated environment, utilizing the first two fidelity levels. 
Then, we incorporate the third fidelity level to apply the algorithm to a real-world environment, effectively refining and adapting the policy for flight experiments.

\subsection{Implementation Details}

The input of the planning policy is normalized as $x^{\text{in}}_{\text{policy}, i} = [\tilde{p}^i_{r}/(L_{\text{space}}/2) \;\; \cos(\tilde{p}^i_{\psi}) \;\; \sin(\tilde{p}^i_{\psi}) \;\; f_{\text{EOS}}]$, where $f_{\text{EOS}}$ is a token indicating the end of the sequence.
Following normalization, the input data undergoes embedding into a 64-dimensional vector through a 2-layer MLP, which is then utilized as the encoder input.
The encoder's initial hidden state is derived from the normalized current trajectory state $\begin{bmatrix}p_r^{(1)}/(L_{\text{space}}/2) \kern2pt \cdots \kern2pt p_r^{(4)}/(L_{\text{space}}/2) \kern2pt p_\psi^{(1)} \kern2pt p_\psi^{(2)} \end{bmatrix}$, passed through a Sigmoid function, and further processed by a 3-layer MLP.
The VAE module compresses the last hidden state of the encoder into a 64-dimensional low-dimensional vector via a 2-layer MLP and then reconstructs it as the initial hidden layer for the decoder using another 2-layer MLP.
The decoder's output is subsequently processed using 3-layer MLPs dedicated to each fidelity level, generating reconstructed waypoints, latent time allocation, and smoothness weight separately for each level.

In the reward estimator, the input includes time allocation and smoothness weights in addition to the waypoint sequences: $x^{\text{in}}_{\text{reward}, i} = [\tilde{p}^i_{r}/(L_{\text{space}}/2) \;\; \cos(\tilde{p}^i_{\psi}) \;\; \sin(\tilde{p}^i_{\psi}) \;\; f_{\text{EOS}} \;\; x_{t,i}/T_{avg} \;\; x_{w,i}]$.
Similar to the planning policy, the input data is embedded into a 64-dimensional vector, with initialization of hidden states from the normalized current state and the use of identical MLP sizes. 
The final hidden state of the GRU is converted into a 64-dimensional feature vector via a 2-layer MLP. 
The MFGPC module predicts feasibility from this feature vector, Gaussian process is approximated with 128 inducing points.
The regularization coefficient between fidelity levels, denoted as $c_{\text{reg}}$ in \eqref{eqn:alg_modeling_mfdgp_loss_all}, is set to 1e-4.
Across the encoder, decoder, and all MLPs, the hidden layer dimensions are consistently set to 256.
The reward parameters are configured as follows: $r_{max} = 0.1$, $r_{bias} = 0.1$, the reward decay factor is $\gamma = 0.9$, and the PPO clipping parameter is $\epsilon = 0.2$.
Both policy and reward estimator updates utilize the Adam optimizer with a learning rate of \num{1e-4}.

% The proposed algorithm is then applied to real-world scenarios by incorporating the third fidelity level into the training, demonstrating the method's ability to efficiently refine and adapt the policy for real-world flight experiments.
The training dataset $\mathcal{D}_{init}$ is made up of $10^5$ sequences of five to fourteen waypoints scaled to the size of the test rooms ($L_{\text{space}}=[9\si{\metre}, 9\si{\metre}, 3\si{\metre}],\:\text{and}\; [20\si{\metre}, 20\si{\metre}, 4\si{\metre}]$).
The parameters for minimum snap optimization in \eqref{eqn:naiveminsnap} are determined through grid search as follows: $\mu_r = 1$, $\mu_{\psi} = 1$, and $\rho = 0$, which are identified to generate the fastest trajectories on average within our simulated evaluation setup.
Table \ref{tab:wp_data} demonstrates the characteristics of waypoint sequences in the generated dataset.

\begin{table}[h!]
% \tiny
\footnotesize
\centering
\caption{
Average distance and curvature between consecutive waypoints in the training dataset, with pairwise traversal times obtained by the snap minimization method.
}
% \vspace{-0.5em}
\label{tab:wp_data}
\begin{tabular}{c ccc}
\toprule
\multicolumn{1}{c}{$L_{\text{space}}$} &
\multicolumn{1}{c}{Distance} &
\multicolumn{1}{c}{Curvature} &
\multicolumn{1}{c}{Time} \\ 
\midrule
$[9 \si{\metre}, 9 \si{\metre}, 3 \si{\metre}]$ & 4.46 $\si{\metre}$ & 0.25 $\si{\metre}^{-1}$ & 1.31 $\si{s}$ \\
$[20 \si{\metre}, 20 \si{\metre}, 4 \si{\metre}]$ & 9.68 $\si{\metre}$ & 0.13 $\si{\metre}^{-1}$ & 1.61 $\si{s}$\\
\hline
\end{tabular}
% \vspace{-0.5em}
\end{table}

The evaluation time varies across fidelity levels, resulting in differing amounts of labeled data for each level. 
Table \ref{tab:exp_num_eval} highlights the differences in evaluation time across fidelity levels, along with the total number of evaluations conducted at each level during dataset generation and MFRL training.
The multi-fidelity datasets are created by labeling subsets of the lower-fidelity dataset with the corresponding evaluation method.
All training datasets are augmented eightfold by rotating and flipping the sequences.
To determine the trajectory time at each fidelity level using the line search procedure described in \eqref{eqn:minsnap_linesearch}, a binary search method is employed, utilizing 10 evaluations for each sequence.
During a single epoch, the planning policy is updated with 800 batches, consisting of 200 waypoint sequences.
Within each batch, we evaluate 16 samples by checking reference motor commands and 1 sample using simulation. 
In the real-world scenarios, we additionally select 100 sequences every 100 epochs and label the output with the real-world experiments.
Training is conducted in both simulated and real-world scenarios until convergence, which takes 1,800 epoch in both cases; 1,800 trajectories are thus evaluated with the real-world flight experiments.

\begin{table}[h!]
% \tiny
\footnotesize
\centering
\caption{
Comparison of evaluation times and evaluation counts across fidelity levels. 
\textit{$\mathcal P_1$} (ideal dynamics), \textit{$\mathcal P_2$} (simulation), and \textit{$\mathcal P_3$} (real-world experiments) represent each fidelity level. 
\textit{Evaluation time} is the average per-trajectory evaluation time. 
\textit{Dataset size} indicates labeled training sequences. 
\textit{Per-epoch BO subset} shows the number of samples selected for reward estimator optimization.
\textit{Number of evaluations} represents the total evaluations during dataset generation and MFRL training.
}
% \vspace{-0.5em}
\label{tab:exp_num_eval}
\begin{tabular}{l ccc}
\toprule
\multicolumn{1}{c}{} &
\multicolumn{1}{c}{$\mathcal P_1$} &
\multicolumn{1}{c}{$\mathcal P_2$} &
\multicolumn{1}{c}{$\mathcal P_3$} \\ 
\midrule
Evaluation time & 75.35 \si{ms} & 2.30 \si{s} & $\sim$ 2 \si{min} \\
% Dataset & 1,000,000 & 20,000 & 500 \\
% RL & 23,040,000 & 1,440,000 & 1,800 \\
\midrule
Dataset size & $|\mathcal{D}_{1}|$\,=\,100,000 & $|\mathcal{D}_{2}|$\,=\,2,000 & $|\mathcal{D}_{3}|$\,=\,50 \\
\midrule
Per-epoch BO subset & 12,800 & 200 & 1 \\
\midrule
\midrule
\multicolumn{2}{l}{Number of evaluations} \\ 
\midrule
Dataset generation & 1,000,000 & 20,000 & 500 \\
MFRL training & 23,040,000 & 1,440,000 & 1,800 \\
\hline
\end{tabular}
% \vspace{-0.5em}
\end{table}

To enhance training efficiency, evaluations for the first and second fidelity levels are processed on an external CPU server with 28 Intel Xeon nodes and 72 cores, parallel to policy updates on an Nvidia RTX 6000 ADA GPU.
Multiple trajectory evaluations are executed concurrently across available cores. 
The second fidelity level, including full controllers and dynamics simulation, demands significantly more memory compared to the first fidelity level, which limits the number of concurrent evaluations possible. 
The majority of training time, approximately 28 hours per epoch, is consumed by these evaluations, particularly those at the second fidelity level.
In real-world scenarios, flight experiments are carried out every 100 epochs, evaluating 100 trajectories over 4 hours. 
The total training procedures across both scenarios are completed over 3 weeks, mostly due to the multi-fidelity evaluation to refine the reward estimator, while RL updates are conducted in parallel without becoming a bottleneck. 
Training dataset generation takes a few days as it requires a much smaller dataset, estimating the same trajectory multiple times at varying velocities.

\begin{table*}[h!]
% \tiny
\footnotesize
\centering
\caption{Comparison of tracking error and trajectory time between the minimum-snap trajectories (\textit{MinSnap}) and trajectories generated by the trained policy (\textit{MFRL}). The policy is trained with the first and second fidelity levels in the space size $L_{\text{space}}=[9\si{\metre}, 9\si{\metre}, 3\si{\metre}]$. Waypoints undergo random shifts within the range of 0 to 3 meters (\textit{Deviation - Pos}) and rotations of 0 to 45 degrees (\textit{Deviation - Yaw}). \textit{Pos error} and \textit{Yaw error} denote mean and standard deviation of position ($\bar{r}^\text{err}$) and yaw ($\bar{\psi}^\text{err}$) tracking errors.
\textit{$\Delta$ Time} represents the trajectory time reduction achieved with the trained policy model relative to the minimum-snap trajectories.  A positive time reduction indicates that the trained policy results in less trajectory time; for instance, MFRL produces trajectories that are \textbf{4.70\%} faster than the baseline when waypoints are unchanged.}
% \vspace{-0.5em}
\label{tab:exp_alg_res_sim}
\begin{tabular}{cccccc}
\toprule
\multicolumn{1}{c}{Deviation} & \multicolumn{2}{c}{MinSnap} & \multicolumn{3}{c}{MFRL} \\ 
\cmidrule(lr){1-1}\cmidrule(lr){2-3}\cmidrule(lr){4-6}
\multicolumn{1}{c}{Pos / Yaw} &
\multicolumn{1}{c}{Pos error} &
\multicolumn{1}{c}{Yaw error} &
\multicolumn{1}{c}{Pos error} &
\multicolumn{1}{c}{Yaw error} &
\multicolumn{1}{c}{$\Delta$ Time} \\ 
\midrule
0.0~$\si{\metre}$ /\hfill\hspace{0.2em} 0~$\si{\degree}$  & 
0.08 $\pm$ 0.02~$\si{\metre}$ & 12.8 $\pm$ 1.3~$\si{\degree}$  & 
0.09 $\pm$ 0.01~$\si{\metre}$ & 12.0 $\pm$ 3.7~$\si{\degree}$  & 4.70 \% \\

1.0~$\si{\metre}$ / 15~$\si{\degree}$ & 
0.16 $\pm$ 2.08~$\si{\metre}$ & 21.5 $\pm$ 29.4~$\si{\degree}$ & 
0.09 $\pm$ 0.03~$\si{\metre}$ & 13.5 $\pm$ 10.0~$\si{\degree}$  & 4.50 \% \\

2.0~$\si{\metre}$ / 30~$\si{\degree}$ & 
0.26 $\pm$ 2.91~$\si{\metre}$ & 40.1 $\pm$ 52.7~$\si{\degree}$ & 
0.11 $\pm$ 0.07~$\si{\metre}$ & 19.2 $\pm$ 25.1~$\si{\degree}$ & 3.51 \% \\

3.0~$\si{\metre}$ / 45~$\si{\degree}$ & 
0.36 $\pm$ 3.09~$\si{\metre}$ & 60.7 $\pm$ 66.7~$\si{\degree}$ & 
0.13 $\pm$ 0.14~$\si{\metre}$ & 27.3 $\pm$ 39.2~$\si{\degree}$ & 2.49 \% \\
\hline
\end{tabular}
% \vspace{-0.5em}
\end{table*}

\subsection{Evaluation in Simulated Environment}

\begin{figure*}[!h]
\begin{subfigure}[b]{0.245\textwidth}
    \captionsetup{justification=centering}
    \includegraphics[width=\textwidth,trim=0.cm 0.cm 0.cm 0.cm,clip]{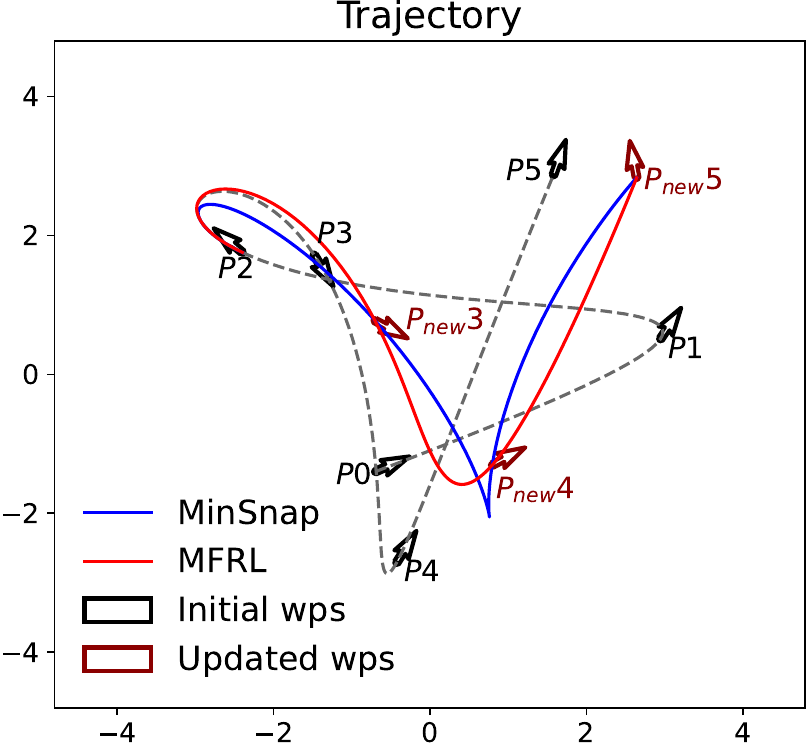}
    \vspace{-1.3\baselineskip}
    \caption{Initial and updated waypoints}
    \label{fig:sim_res_dev_ms_usage_traj}
\end{subfigure}
\begin{subfigure}[b]{0.3675\textwidth}
    \captionsetup{justification=centering}
    \includegraphics[width=\textwidth,trim=0.cm 0.cm 0.cm 0.cm,clip]{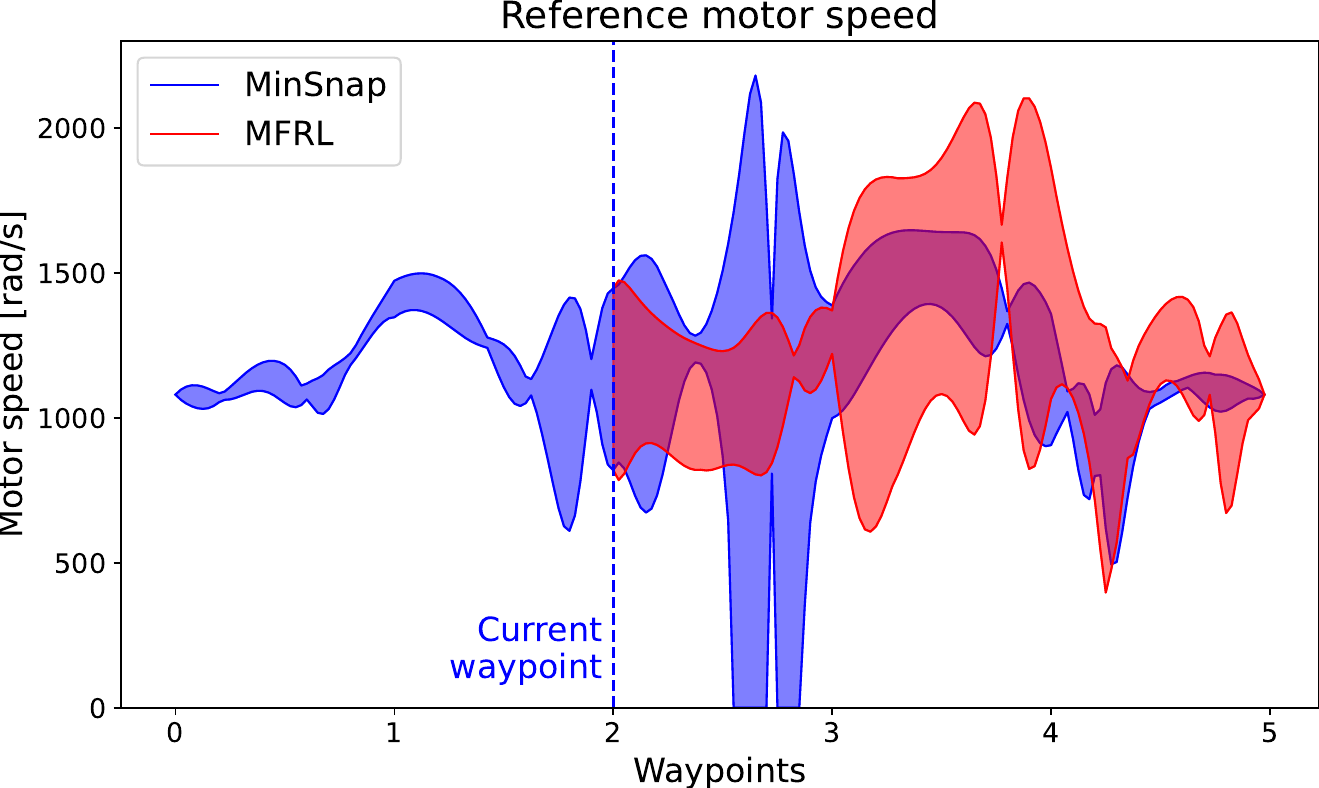}
    \vspace{-1.3\baselineskip}
    \caption{Reference motor speeds over waypoints}
    \label{fig:sim_res_dev_ms_usage_sta}
\end{subfigure}
\begin{subfigure}[b]{0.3675\textwidth}
    \captionsetup{justification=centering}
    \includegraphics[width=\textwidth,trim=0.cm 0.cm 0.cm 0.cm,clip]{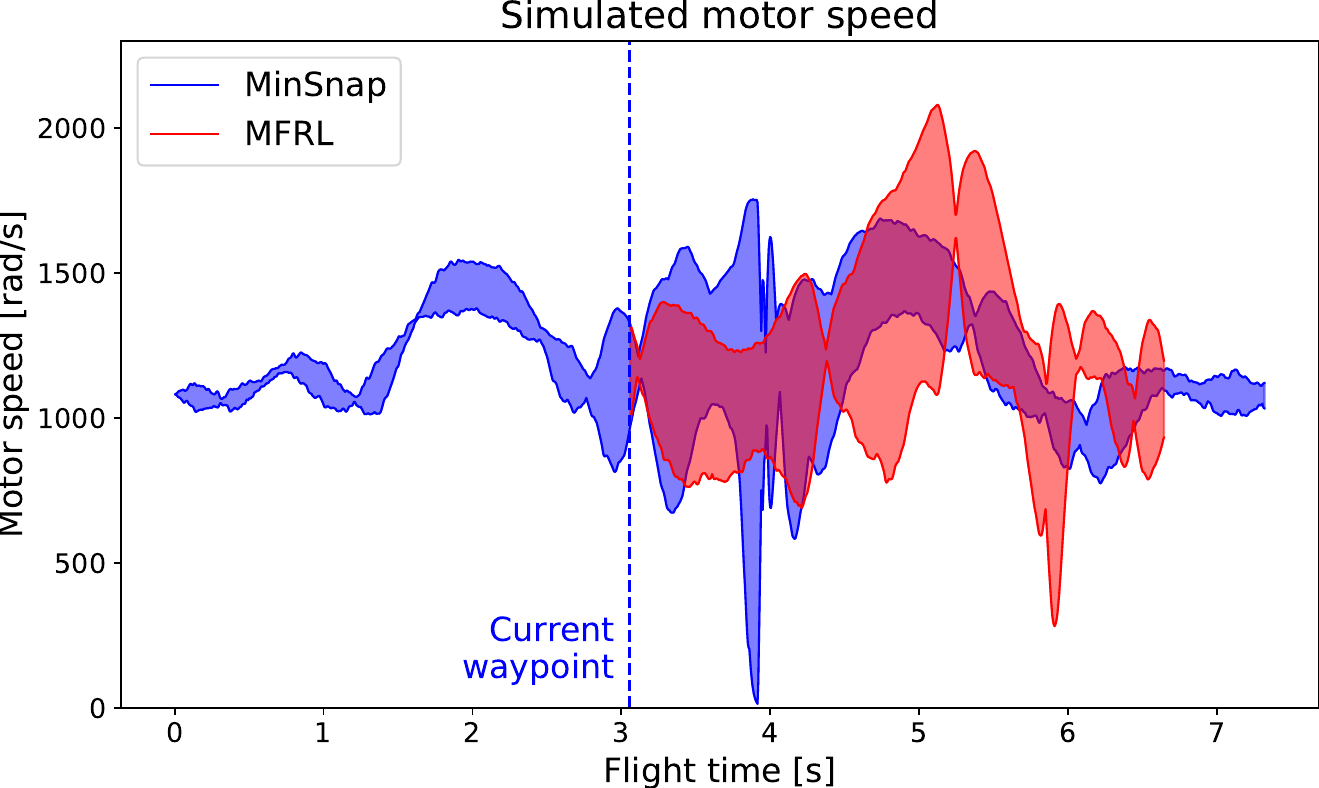}
    \vspace{-1.3\baselineskip}
    \caption{Simulated motor speeds over flight time}
    \label{fig:sim_res_dev_ms_usage_sim}
\end{subfigure}
\vspace{.3\baselineskip}

\begin{subfigure}[b]{0.49\textwidth}
    \captionsetup{justification=centering}
    \includegraphics[width=\textwidth,trim=0.cm 0.cm 0.cm 0.cm,clip]{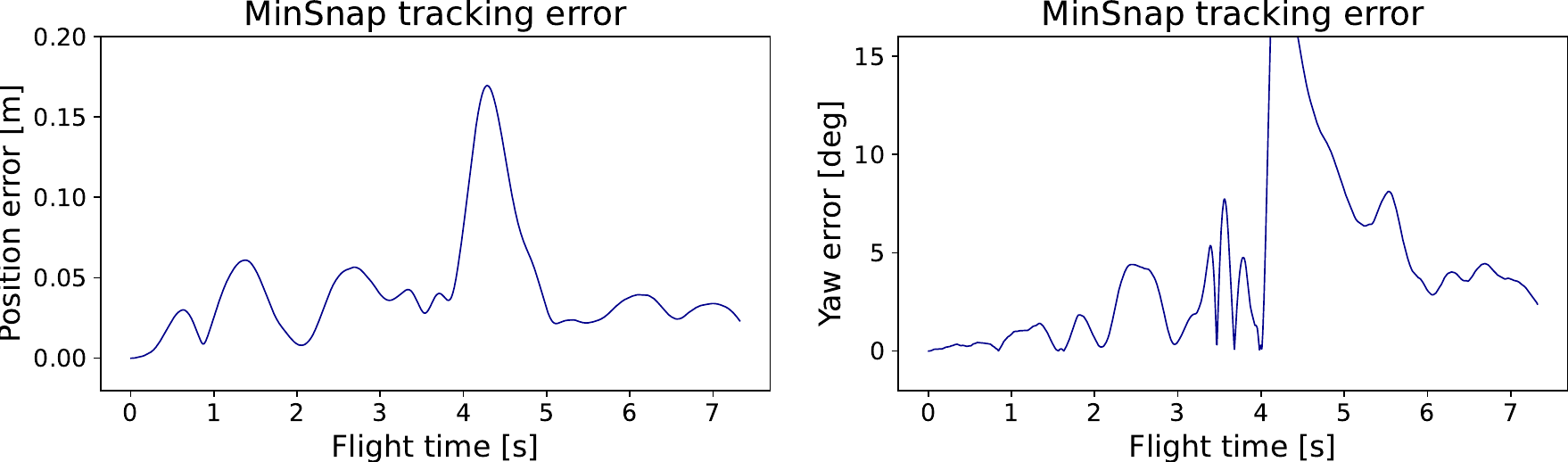}
    \vspace{-1.3\baselineskip}
    \caption{Tracking error using the initial time allocation}
    \label{fig:sim_res_dev_ms_usage_ms}
\end{subfigure}
\begin{subfigure}[b]{0.49\textwidth}
    \captionsetup{justification=centering}
    \includegraphics[width=\textwidth,trim=0.cm 0.cm 0.cm 0.cm,clip]{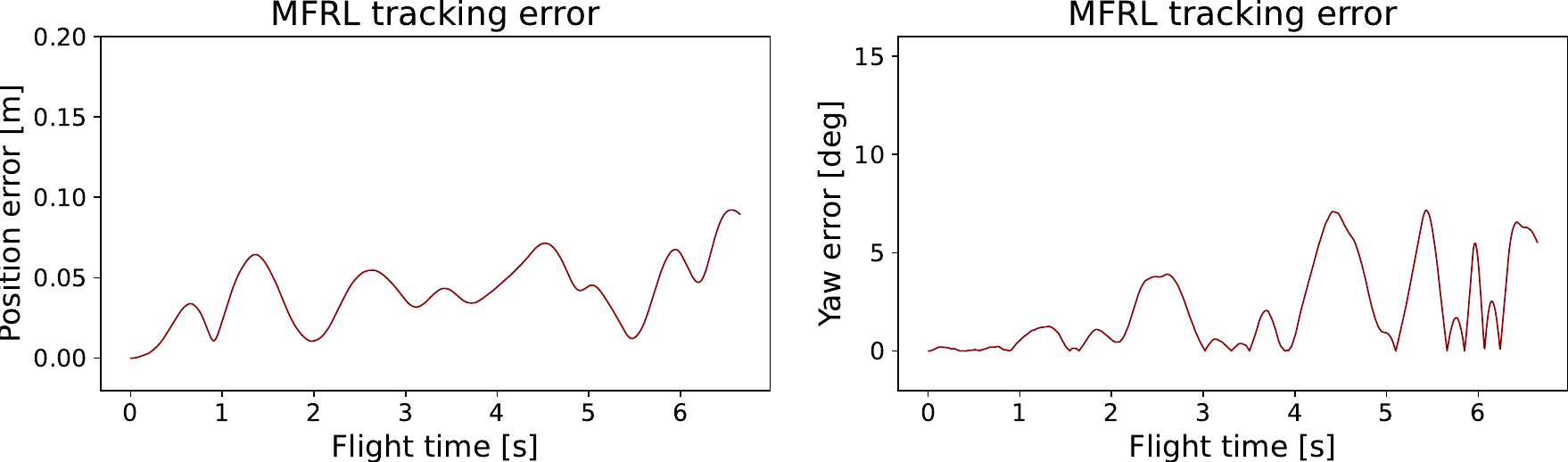}
    \vspace{-1.3\baselineskip}
    \caption{Tracking error using the adapted time allocation}
    \label{fig:sim_res_dev_ms_usage_rl}
\end{subfigure}
\vspace{.3\baselineskip}

\begin{subfigure}[b]{0.49\textwidth}
    \captionsetup{justification=centering}
    \includegraphics[width=\textwidth,trim=0.cm 0.cm 0.cm 0.cm,clip]{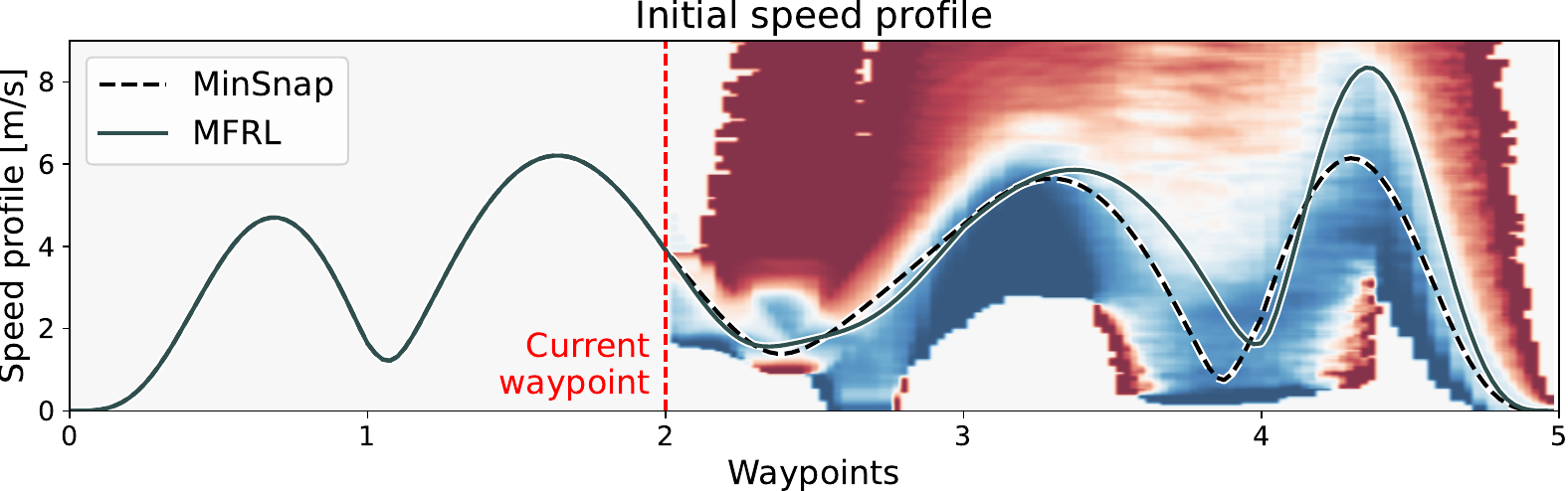}
    \vspace{-1.3\baselineskip}
    \caption{Comparison of the initial speed profile}
    \label{fig:sim_res_dev_ms_usage_spd_i}
\end{subfigure}
\begin{subfigure}[b]{0.49\textwidth}
    \captionsetup{justification=centering}
    \includegraphics[width=\textwidth,trim=0.cm 0.cm 0.cm 0.cm,clip]{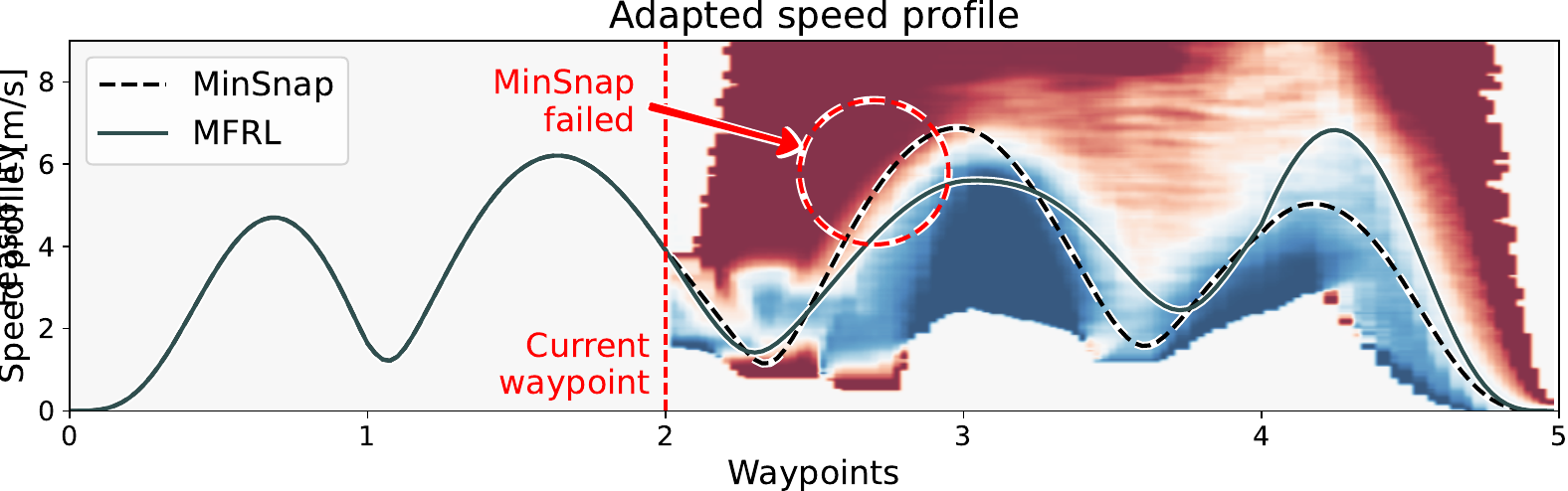}
    \vspace{-1.3\baselineskip}
    \caption{Comparison of the adapted speed profile}
    \label{fig:sim_res_dev_ms_usage_spd_f}
\end{subfigure}
\vspace{-.5\baselineskip}

\caption{
Comparison between minimum snap and adapted trajectories in response to waypoint deviations, where waypoints are shifted by 2 meters and rotated by 30 degrees.
(a) The initial and updated waypoints are shown alongside baseline and trained policy trajectories.
(b) Reference motor speed commands equally sampled from each trajectory segment. 
(c) Simulated motor speeds, demonstrating that the adapted trajectory is 9.22\% faster than the baseline while maintaining motor speeds within the admissible range. 
(d) and (e) Comparison of tracking errors, where the adapted trajectory remains below the threshold, unlike the minimum snap trajectory.
The last row, (f) and (g) presents speed profiles before and after waypoint deviations.
Background feasibility colormap is generated by random speed profile sampling and averaging of reference motor speed feasibility.
Both trajectories are initially within the feasible region (f), but as waypoints change, the MFRL policy keeps the trajectory feasible, while the MinSnap baseline ventures into infeasible regions (g).
}
\label{fig:sim_res_dev_ms_usage}
% \vspace{-1.0em}
\end{figure*}

To evaluate our policy, we reconstruct trajectories starting from the midpoint of the waypoint sequences. 
Subsequently, we introduce random shifts and rotations to the remaining waypoints to assess whether our policy could generate feasible trajectories. 
The testing dataset is derived from 4,000 randomly generated waypoint sequences, with trajectory times ($T^{\text{MS}}$) determined through numerical simulations. 
The testing trajectories consist of an average speed of 3.9 $\si{m/s}$, with a maximum of 11.2 $\si{m/s}$, and an average acceleration of 0.7 g, with a maximum of 2.7 g, where g denotes standard gravity acceleration ($\text{g}=9.81 \si{m/s^2}$).

Table \ref{tab:exp_alg_res_sim} presents a comparison between the baseline minimum snap (MinSnap) trajectory and trajectories generated by our trained policy (MFRL). 
Along with the time reduction, we compare the mean and standard deviation of tracking errors for trajectories.
The position and yaw tracking errors, $\bar{r}^\text{err}$ and $\bar{\psi}^\text{err}$, are calculated as the maximum deviation between the vehicle's pose and the reference pose:
\begin{align}
\bar{r}^\text{err} &= \max_t\:\lVert p_r(t) - r(t) \rVert, \nonumber \\
\bar{\psi}^\text{err} &= \max_t\:|p_\psi(t) - \psi(t)|
\end{align}
where $r(t)$ and $\psi(t)$ are the reference position and yaw angle respectively, as defined in \eqref{eqn:feasibility_tracking_err}. 
$p_r(t)$ and $p_\psi(t)$ represent the simulated position and yaw angle of the vehicle at time $t$.
For clarity, while constraining the tracking error maintains a marginal time difference between actual and reference trajectory times, this difference is negligible compared to our method's overall time reduction.
When there are deviations in the position and yaw of the remaining waypoints, the baseline technique employs the original trajectory's time allocation and solves the minimum-snap quadratic programming problem using the updated waypoint positions to adapt the trajectory.
As the remaining waypoints deviate further, the baseline method encounters greater average tracking error as well as increased error variance, reflecting its struggle to track a larger portion of the trajectory.
On the other hand, the trained policy adjusts the time allocations to generate a trajectory that consistently maintains acceptable tracking errors.
The table also showcases the time reduction ($\Delta$ Time) achieved by our trained policy, where a positive reduction indicates faster trajectories compared to the baseline method. 
Initial time allocations are optimized for smooth trajectories; thus, adapting these allocations demands extra time in response to random waypoint deviations.
Consequently, time reduction is less in such scenarios compared to non-deviation cases. 
Even so, the trained model consistently achieves faster trajectories than the baseline.
Figure \ref{fig:sim_res_dev_ms_usage} compares motor speed and tracking error between the baseline method and the trained policy on an example trajectory with randomly deviated waypoints. 
As a response to these deviations, the trained policy updates the trajectory to maintain an admissible tracking error and reduce flight time, unlike the baseline method, which fails to generate a feasible trajectory.

% Heuristic sqrt
\begin{table*}[h!]
% \scriptsize
\footnotesize
\centering
\caption{Tracking error and trajectory time of the trajectory adapted with the heuristic method (\textit{MinSnap Heuristic}) and the trajectory obtained from the model only trained with supervised learning (\textit{Supervised-only}). 
Both the heuristic method and the supervised-only trained model yield slower trajectories and higher tracking errors compared to the MFRL policy.}
% \vspace{-0.5em}
\label{tab:exp_alg_res_sim_comp}
\begin{tabular}{c ccc ccc}
\toprule
\multicolumn{1}{c}{Deviation} & \multicolumn{3}{c}{MinSnap Heuristic} & \multicolumn{3}{c}{Supervised-only} \\ 
\cmidrule(lr){1-1}\cmidrule(lr){2-4}\cmidrule(lr){5-7}
\multicolumn{1}{c}{Pos / Yaw} &
\multicolumn{1}{c}{Pos error} &
\multicolumn{1}{c}{Yaw error} &
\multicolumn{1}{c}{$\Delta$ Time} &
\multicolumn{1}{c}{Pos error} &
\multicolumn{1}{c}{Yaw error} &
\multicolumn{1}{c}{$\Delta$ Time} \\ 
\midrule
0.0~$\si{\metre}$ /\hfill\hspace{0.2em} 0~$\si{\degree}$  & 
0.08 $\pm$ 0.02~$\si{\metre}$ & 12.9 $\pm$ 1.3~$\si{\degree}$  & 0.00 \%  & 
0.08 $\pm$ 0.01~$\si{\metre}$ & 13.2 $\pm$ 4.7~$\si{\degree}$  & -0.69 \% \\

1.0~$\si{\metre}$ / 15~$\si{\degree}$ & 
0.17 $\pm$ 2.36~$\si{\metre}$ & 21.0 $\pm$ 27.9~$\si{\degree}$ & -0.50 \%  & 
0.09 $\pm$ 0.04~$\si{\metre}$ & 16.0 $\pm$ 16.6~$\si{\degree}$ & -1.15 \% \\

2.0~$\si{\metre}$ / 30~$\si{\degree}$ & 
0.24 $\pm$ 2.39~$\si{\metre}$ & 39.7 $\pm$ 52.2~$\si{\degree}$ & -1.34 \%  & 
0.11 $\pm$ 0.09~$\si{\metre}$ & 24.1 $\pm$ 34.7~$\si{\degree}$ & -2.01 \% \\

3.0~$\si{\metre}$ / 45~$\si{\degree}$ & 
0.39 $\pm$ 3.50~$\si{\metre}$ & 58.3 $\pm$ 65.3~$\si{\degree}$ & -2.14 \% & 
0.13 $\pm$ 0.18~$\si{\metre}$ & 32.5 $\pm$ 46.2~$\si{\degree}$ & -3.12 \% \\
\hline
\end{tabular}
% \vspace{-0.5em}
\end{table*}

\begin{table*}[h!]
% \scriptsize
\footnotesize
\centering
\caption{Comparison of tracking error and trajectory time between the minimum-snap trajectories (\textit{MinSnap}) and trajectories generated by the trained policy (\textit{MFRL}). The policy is trained with the first and second fidelity levels in the large target space $L_{\text{space}}=[20\si{\metre}, 20\si{\metre}, 4\si{\metre}]$. The trained policy generates trajectories that are 6.23\% faster than the baseline when waypoints are unchanged.}
% \vspace{-0.5em}
\label{tab:exp_alg_res_sim_large}
\begin{tabular}{cccccc}
\toprule
\multicolumn{1}{c}{Deviation} & \multicolumn{2}{c}{MinSnap} & \multicolumn{3}{c}{MFRL} \\ 
\cmidrule(lr){1-1}\cmidrule(lr){2-3}\cmidrule(lr){4-6}
\multicolumn{1}{c}{Pos / Yaw} &
\multicolumn{1}{c}{Pos error} &
\multicolumn{1}{c}{Yaw error} &
\multicolumn{1}{c}{Pos error} &
\multicolumn{1}{c}{Yaw error} &
\multicolumn{1}{c}{$\Delta$ Time} \\ 
\midrule
0.0~$\si{\metre}$ /\hfill\hspace{0.2em} 0~$\si{\degree}$  & 
0.09 $\pm$ 0.02~$\si{\metre}$ & 13.4 $\pm$ 0.8~$\si{\degree}$  & 
0.12 $\pm$ 0.56~$\si{\metre}$ & 12.9 $\pm$ 7.0~$\si{\degree}$  & 6.23 \% \\

2.0~$\si{\metre}$ / 15~$\si{\degree}$ & 
0.14 $\pm$ 1.30~$\si{\metre}$ & 18.5 $\pm$ 21.4~$\si{\degree}$ & 
0.13 $\pm$ 0.63~$\si{\metre}$ & 13.7 $\pm$ 8.8~$\si{\degree}$  & 5.74 \% \\

4.0~$\si{\metre}$ / 30~$\si{\degree}$ & 
0.24 $\pm$ 2.16~$\si{\metre}$ & 29.3 $\pm$ 37.5~$\si{\degree}$ & 
0.14 $\pm$ 0.87~$\si{\metre}$ & 16.8 $\pm$ 17.8~$\si{\degree}$ & 4.71 \% \\

6.0~$\si{\metre}$ / 45~$\si{\degree}$ & 
0.46 $\pm$ 3.24~$\si{\metre}$ & 42.0 $\pm$ 50.0~$\si{\degree}$ & 
0.15 $\pm$ 0.54~$\si{\metre}$ & 20.3 $\pm$ 25.0~$\si{\degree}$ & 4.49 \% \\
\hline
\end{tabular}
% \vspace{-0.5em}
\end{table*}

The incorporation of the first fidelity level evaluations proves essential to our methodology. 
Their absence would necessitate a prohibitively large amount of the second fidelity evaluations to replace the current data volume, making collection impractically time-consuming.
Alternatively, maintaining only the current limited quantity of the second fidelity evaluations without the first fidelity data would result in highly inaccurate feasibility constraint predictions, causing the MFRL training process to diverge.

In Table \ref{tab:exp_alg_res_sim_comp}, we compare our approach with two widely used methods: 1) MinSnap Heuristic: a heuristic method that adjusts time allocation proportional to the trajectory length, chosen for its comparable computation speed:
\begin{equation}
x_{t,i} = \frac{\lVert \tilde p_\text{new}^i - \tilde p_\text{new}^{i-1} \rVert}{\lVert \tilde p^i - \tilde p^{i-1} \rVert}\: x^\text{MS}_{t,i}
\end{equation}
and 2) Supervised-only: a policy model trained to imitate minimum snap optimization output, serving as a proxy for comparison with optimization-based methods.
The pretrained model, utilized for MFRL initialization, is used to evaluate the second method, employing only the time allocation output with the identity smoothness weight.
The first method fails to adapt the trajectory, leading to diverging tracking errors when waypoints are randomly shifted. 
In contrast, the second method can update the trajectory to remain feasible, yet it results in a longer trajectory time compared to the MFRL-trained model.

We further apply the training in a larger space ($L_{\text{space}}=[20\si{\metre}, 20\si{\metre}, 4\si{\metre}]$) using scaled waypoint sequence data. 
In the wider space, trajectories could attain higher linear speeds, resulting in testing trajectories with an average speed of 5.5 $\si{m/s}$ and a maximum of 15.1 $\si{m/s}$, as well as an average acceleration of 0.7 g and a maximum of 2.9 g. 
As shown in Table \ref{tab:exp_alg_res_sim_large}, our trained model effectively updates trajectories while maintaining tracking accuracy, achieving even greater time reduction compared to the smaller space.
In smaller spaces, tracking error bounds are typically determined by yaw tracking error due to the need for agile turns in confined areas. 
In contrast, in larger spaces, position tracking error becomes more prominent, as there is ample room for linear acceleration and reaching the vehicle's maximum speed, providing our MFRL model with additional opportunities for optimization.

\subsection{Analysis of MFRL Policy}
\begin{figure*}[!h]
\centering
\begin{subfigure}[b]{0.24\textwidth}
    \captionsetup{justification=centering}
    \includegraphics[width=\textwidth,trim=0.2cm 0.3cm 0.1cm 0.cm,clip]{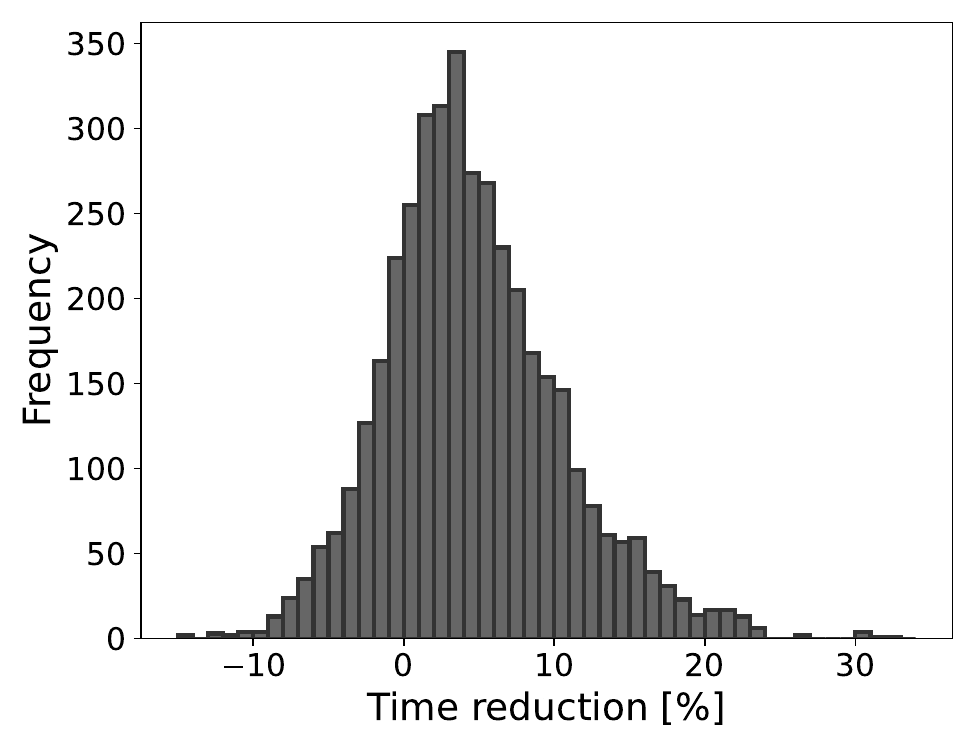}
    \vspace{-1.2\baselineskip}
    \caption{Distribution of time reduction}
    \label{fig:sim_res_imp}
\end{subfigure}
\begin{subfigure}[b]{0.24\textwidth}
    \captionsetup{justification=centering}
    \includegraphics[width=\textwidth,trim=0.2cm 0.3cm 0.1cm 0.cm,clip]{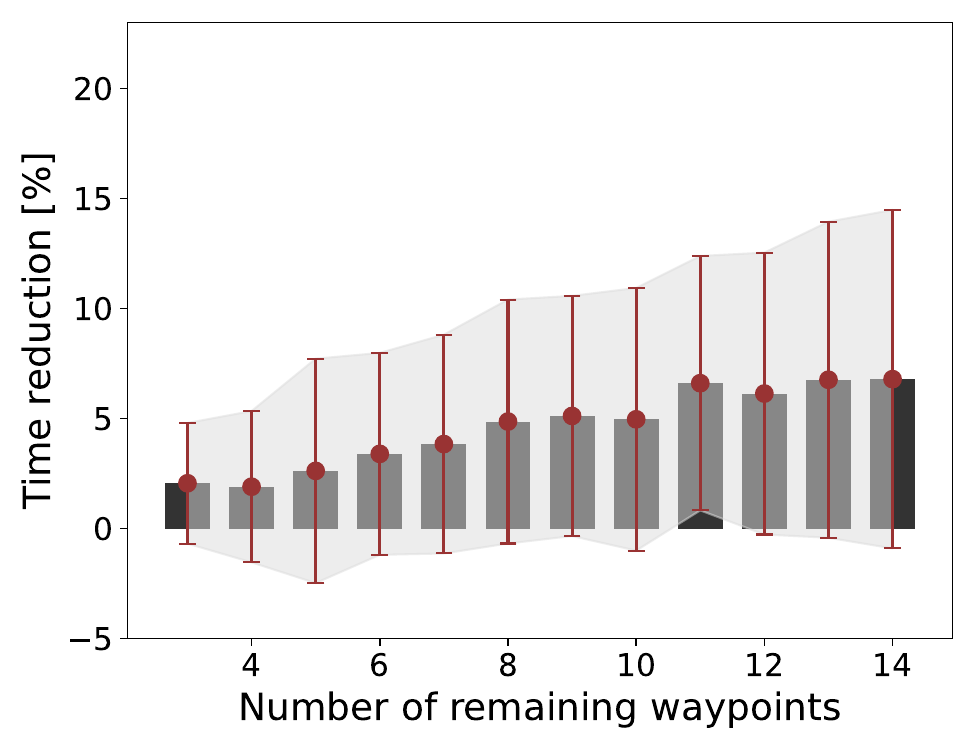}
    \vspace{-1.2\baselineskip}
    \caption{Number of waypoints}
    \label{fig:sim_res_seqlen}
\end{subfigure}
\begin{subfigure}[b]{0.24\textwidth}
    \captionsetup{justification=centering}
    \includegraphics[width=\textwidth,trim=0.2cm 0.3cm 0.1cm 0.cm,clip]{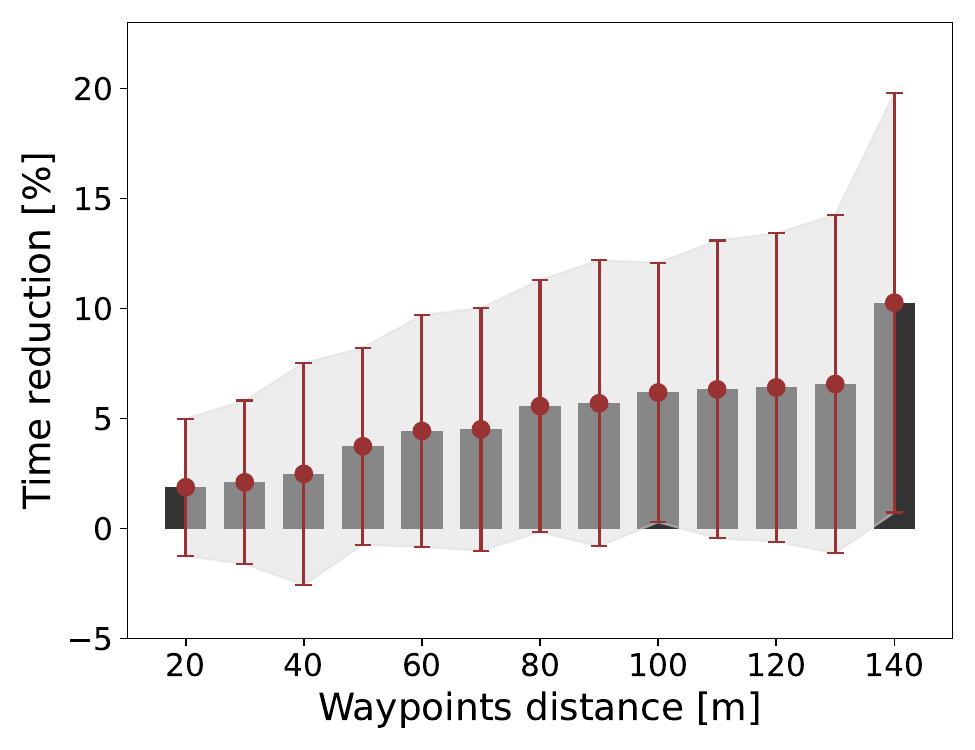}
    \vspace{-1.2\baselineskip}
    \caption{Remaining distance}
    \label{fig:sim_res_wpdist}
\end{subfigure}
\begin{subfigure}[b]{0.24\textwidth}
    \captionsetup{justification=centering}
    \includegraphics[width=\textwidth,trim=0.2cm 0.3cm 0.1cm 0.cm,clip]{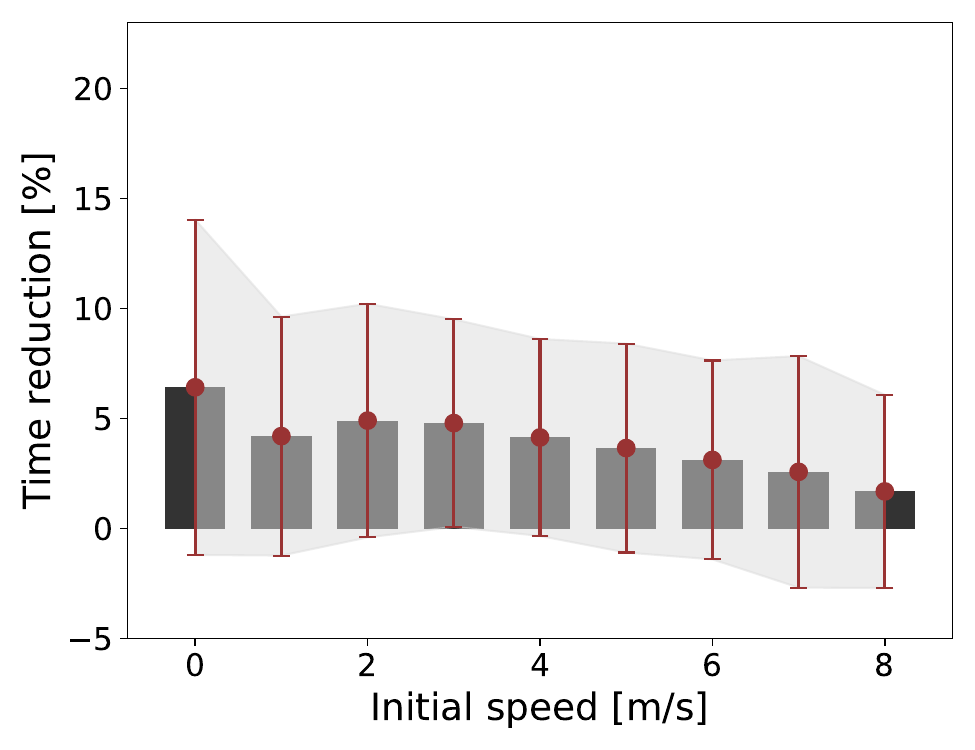}
    \vspace{-1.2\baselineskip}
    \caption{Initial speed}
    \label{fig:sim_res_ispd}
\end{subfigure}
\caption{(a) Histogram depicting the time reduction achieved by the trained policy model in comparison to the minimum-snap trajectories. (b) Time reduction regarding the number of remaining waypoints. (c) Time reduction regarding the distance between the remaining waypoints. (d) Time reduction regarding the initial speed. For (b-d), the average time reduction is plotted along with the standard deviation.
}
\label{fig:exp_sim_res}
% \vspace{-1.0em}
\end{figure*}

We conduct further analysis of our training results by examining the correlation between trajectory features and time reduction. 
As depicted in the histogram of Figure \ref{fig:sim_res_imp}, the trained policy consistently yields faster trajectories, denoted by positive time reduction, for the majority of waypoint sequences when compared to the baseline method. 
Figure \ref{fig:exp_sim_res} further illustrates the correlation between the number of remaining waypoints, the distance between these waypoints, initial speed, and time reduction. 
Intuitively, trajectories with more degrees of freedom to optimize, such as those with a larger number of remaining waypoints or greater remaining distances, lead the trained policy to generate faster trajectories. 
Likewise, slower initial speeds give the vehicle the flexibility to accelerate more in the remaining trajectory, reducing overall flight time.

A key correlation factor we've identified is motor utilization. 
We define the motor utilization factor $U_{\text{motor}}$ as the average gap between motor speed commands and stationary motor commands, divided by the maximum gap, as described in the following equation:
\begin{equation}
    U_{\text{motor}} = \frac{1}{4N_{\text{sample}}} \sum_{n=1}^{N_{\text{sample}}} \sum_{i=1}^4  \frac{|ms_{n,i} - ms_{\text{sta}}|}{(ms_{\text{max}} - ms_{\text{min}})/2}.
\end{equation}
The maximum gap is approximated as half of the range of admissible motor speed values ($(ms_{\text{max}} - ms_{\text{min}}) / 2$).
This factor is calculated based on equally sampled reference motor speed commands of four rotors $ms_{n,i},\; (i=1,\dots,4)$ from each trajectory segment where $N_{\text{sample}}$ refers to the number of samples.
For instance, when the vehicle follows the trajectory with a bang-bang controller, motor speed commands consistently reach the upper and lower bounds, resulting in a utilization factor close to 100 \%. 

\begin{figure}[!h]
\centering
\begin{subfigure}[b]{0.24\textwidth}
    \includegraphics[width=\textwidth,trim=0.2cm 0.cm 0.1cm 0.cm,clip]{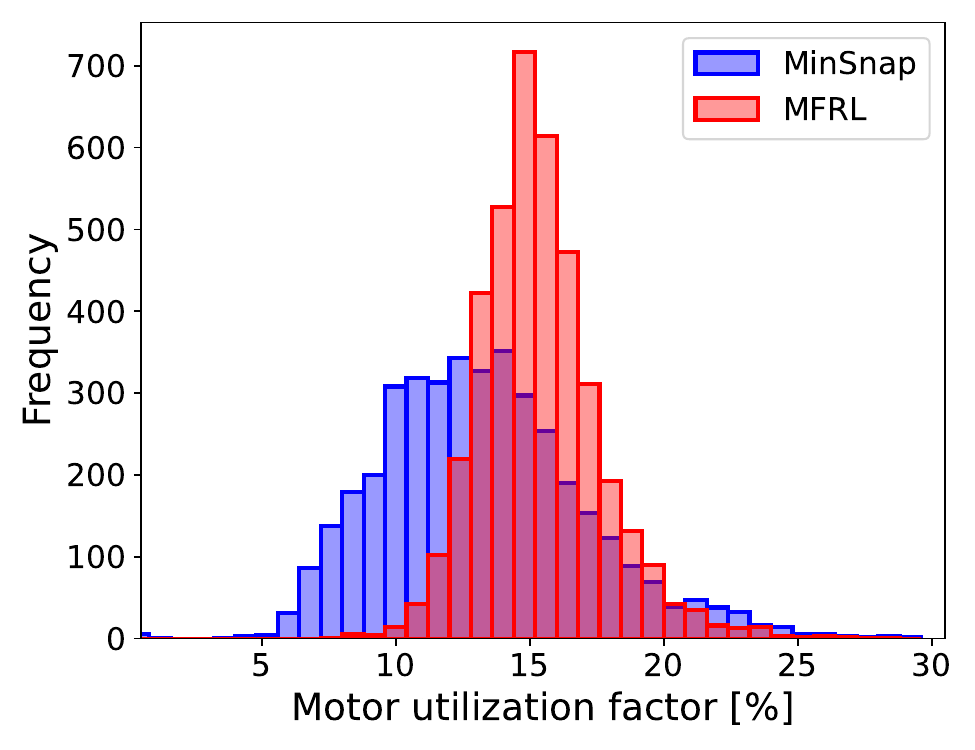}
    \vspace{-1\baselineskip}
    % \caption{Distributions of motor utilization}
    \label{fig:sim_res_motor_usage_hist}
\end{subfigure}
\begin{subfigure}[b]{0.24\textwidth}
    \includegraphics[width=\textwidth,trim=0.2cm 0.cm 0.1cm 0.cm,clip]{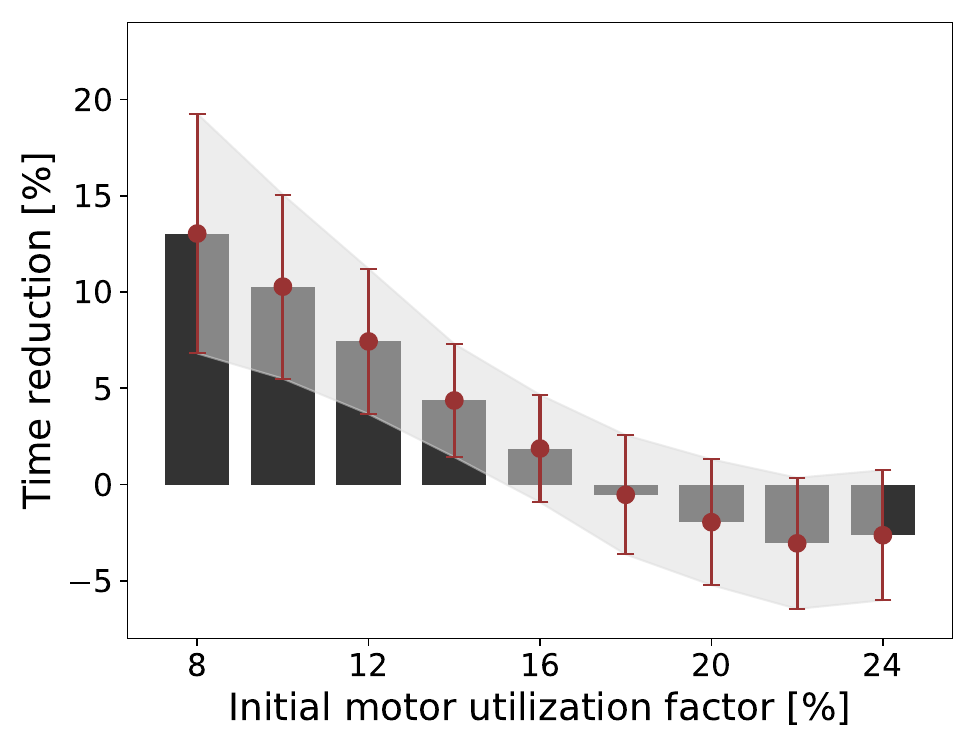}
    \vspace{-1\baselineskip}
    % \caption{Time reduction.}
    \label{fig:sim_res_motor_usage_trend}
\end{subfigure}
\caption{(Left) Histogram comparing motor utilization factors between minimum-snap trajectories (\textit{MinSnap}) and MFRL policy trajectories (\textit{MFRL}). (Right) Time reduction achieved by the trained policy regarding the initial utilization factor of the minimum-snap trajectory. The average time reduction is plotted along with the standard deviation.
}
\label{fig:sim_res_motor_usage}
% \vspace{-1.0em}
\end{figure}

Comparing the utilization factor between baseline trajectories and those from the trained model in Figure \ref{fig:sim_res_motor_usage} reveals that the trained policy reduces the variance of the utilization factor and increases its mean. 
This indicates that MFRL trains the policy to maintain motor utilization at an optimal level, efficiently utilizing the vehicle's capacity while ensuring feasibility. 
As shown in the right plot of Figure \ref{fig:sim_res_motor_usage}, in cases where the initial minimum snap trajectory has low motor utilization, the trained model achieves significantly greater reductions (about 12 \% faster) in flight time. 
Furthermore, the time reduction's variation associated with motor utilization is smaller compared to other trajectory features, highlighting their strong correlation.
This reduced variation demonstrates how the trained policy consistently improves performance based on the remaining optimization margin, as quantified by motor usage.
Table \ref{tab:exp_alg_res_sim_spdacc}, presenting data from the default space size, and Table \ref{tab:exp_alg_res_sim_large_spdacc}, from a larger space, compare parameters such as speed, acceleration, and motor utilization between minimum-snap and MFRL trajectories. 
Both tables reveal that, while average speeds and accelerations are comparable, MFRL trajectories show higher maximums and a significantly greater average motor utilization with lower variance.

\begin{table}[h!]
% \tiny
\footnotesize
\centering
\caption{
Comparison of parameters including speed ($v$), acceleration ($a$), and motor utilization factor ($U_{\text{motor}}$) between minimum-snap trajectories (\textit{MinSnap}) and trajectories generated by the trained policy (\textit{MFRL}) ($L_{\text{space}}=[9\si{\metre}, 9\si{\metre}, 3\si{\metre}]$).}
% \vspace{-0.5em}
\label{tab:exp_alg_res_sim_spdacc}
\begin{tabular}{l ccccc}
\toprule
\multicolumn{1}{c}{} &
\multicolumn{1}{c}{$v_{\text{avg}}$} &
\multicolumn{1}{c}{$v_{\text{max}}$} &
\multicolumn{1}{c}{$a_{\text{avg}}$} &
\multicolumn{1}{c}{$a_{\text{max}}$} &
\multicolumn{1}{c}{$U_{\text{motor}}$} \\ 
\midrule

MinSnap & 3.9 \si{m/s} & 11.2 \si{m/s} & 0.7 \text{g} & 2.7 \text{g} & 13.3 $\pm$ 3.9 \% \\
MFRL & 3.8 \si{m/s} & 12.1 \si{m/s} & 0.8 \text{g} & 3.0 \text{g} & 15.4 $\pm$ 2.3 \% \\

\hline
\end{tabular}
% \vspace{-0.5em}
\end{table}

\begin{table}[h!]
% \tiny
\footnotesize
\centering
\caption{
Comparison of parameters including speed ($v$), acceleration ($a$), and motor utilization factor ($U_{\text{motor}}$) between minimum-snap trajectories (\textit{MinSnap}) and trajectories generated by the trained policy (\textit{MFRL}) ($L_{\text{space}}=[20\si{\metre}, 20\si{\metre}, 4\si{\metre}]$).}
% \vspace{-0.5em}
\label{tab:exp_alg_res_sim_large_spdacc}
\begin{tabular}{l ccccc}
\toprule
\multicolumn{1}{c}{} &
\multicolumn{1}{c}{$v_{\text{avg}}$} &
\multicolumn{1}{c}{$v_{\text{max}}$} &
\multicolumn{1}{c}{$a_{\text{avg}}$} &
\multicolumn{1}{c}{$a_{\text{max}}$} &
\multicolumn{1}{c}{$U_{\text{motor}}$} \\ 
\midrule

MinSnap & 5.5 \si{m/s} & 15.1 \si{m/s} & 0.7 \text{g} & 2.9 \text{g} & 10.0 $\pm$ 3.3 \% \\
MFRL & 5.5 \si{m/s} & 16.3 \si{m/s} & 0.8 \text{g} & 3.3 \text{g} & 13.0 $\pm$ 2.5 \% \\

\hline
\end{tabular}
% \vspace{-0.5em}
\end{table}

\begin{figure*}[!h]
\includegraphics[width=\textwidth,trim=0.cm 0.cm 0.cm 0.cm,clip]{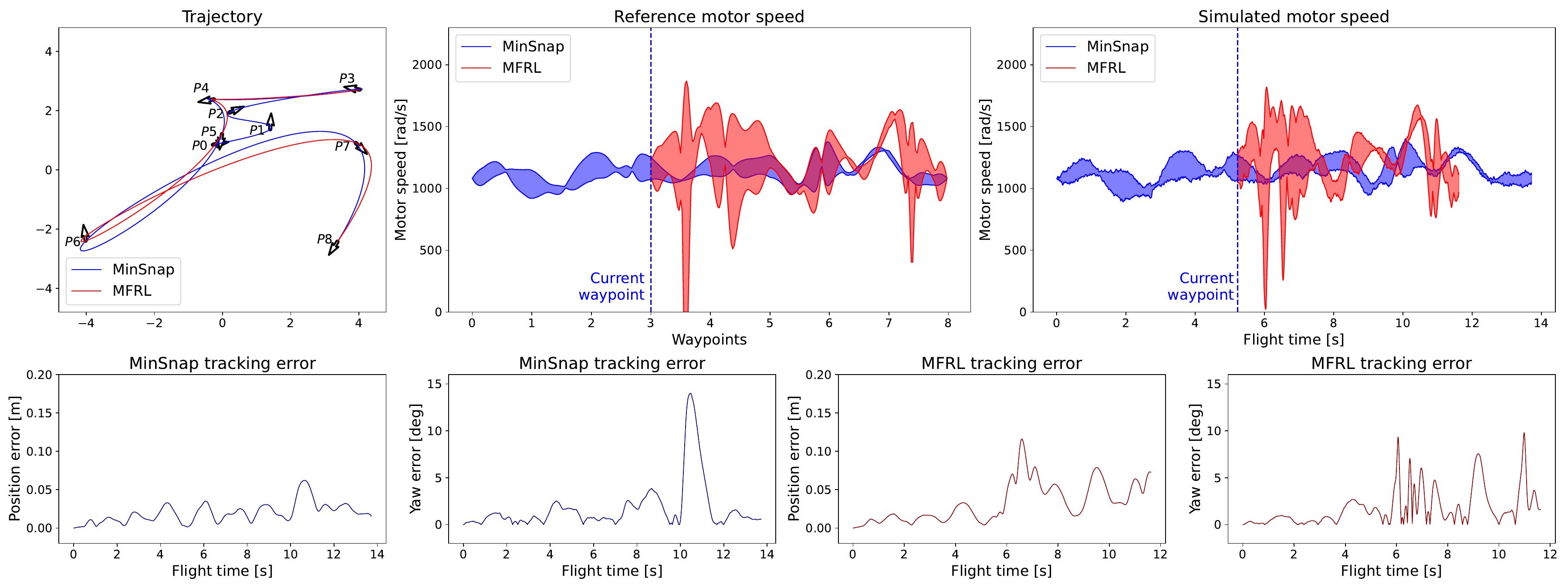}
\vspace{-1\baselineskip}
\caption{
Trajectory with a low initial motor utilization factor of 6.00 \%. This figure compares reference motor commands, simulated motor speed, and tracking error between the initial minimum snap trajectory and the trajectory generated by the trained model. The trained policy reduces the trajectory time by 15.31 \%, concurrently increasing the utilization factor to 16.32 \%.}
\label{fig:sim_res_motor_usage_low}
% \vspace{-1.0em}
\end{figure*}

\begin{figure*}[!h]
\includegraphics[width=\textwidth,trim=0.cm 0.cm 0.cm 0.cm,clip]{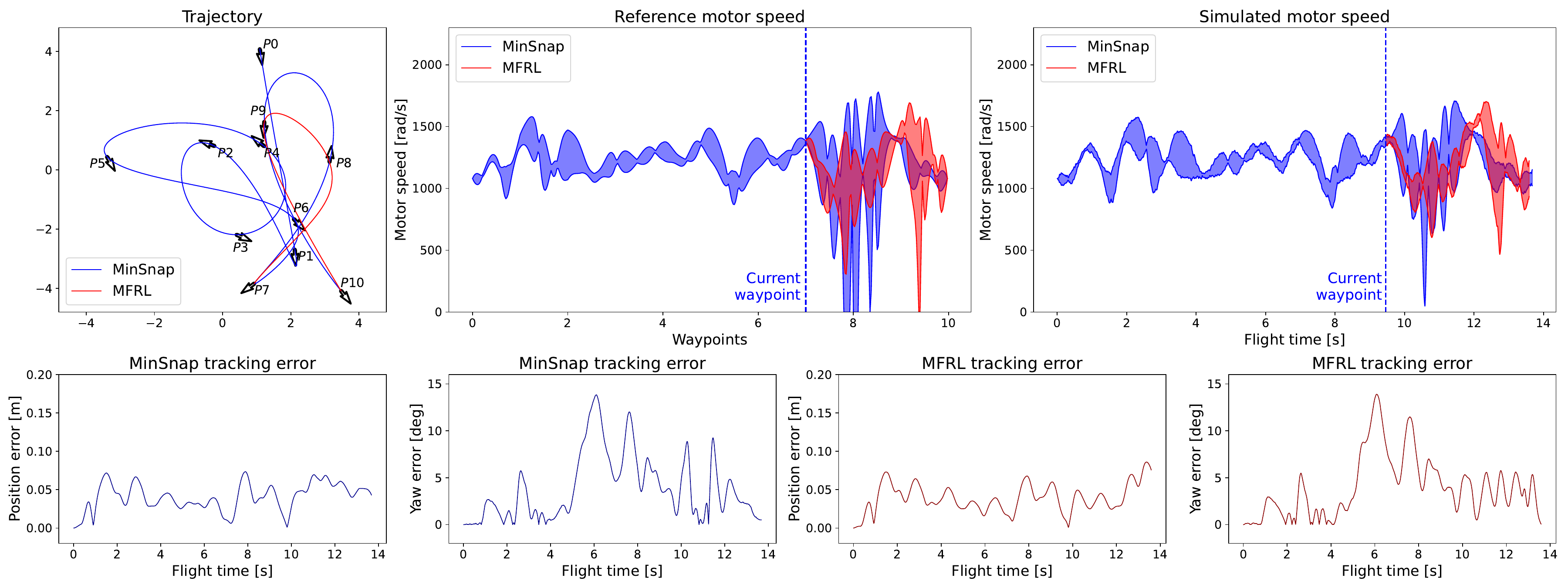}
\vspace{-1\baselineskip}
\caption{
Trajectory with a high initial motor utilization factor of 20.28 \%. This figure compares reference motor commands, simulated motor speed, and tracking error between the initial minimum snap trajectory and the trajectory generated by the trained model. The trained policy reduces the trajectory time by only 0.66 \%, while also decreasing the utilization factor to 16.38 \%.}
\label{fig:sim_res_motor_usage_high}
% \vspace{-1.0em}
\end{figure*}

Figure \ref{fig:sim_res_motor_usage_low} and Figure \ref{fig:sim_res_motor_usage_high} respectively demonstrate scenarios where the initial trajectory has low and high motor utilization factor.
Although the baseline minimum snap trajectory typically generates smoothly changing control commands, the controller may face difficulties in tracking it at high speeds, which is due to factors such as approximations in the dynamics model or the balance of internal control outputs.
% Clarify
The MFRL policy adjusts the trajectory by decreasing motor speeds in segments with high tracking errors and increasing motor speeds in the remaining segments, ultimately reducing the overall flight time.
A high motor utilization factor indicates limited space for trajectory optimization, often due to proximity to the final waypoints or excessively high initial speeds, necessitating immediate deceleration.
In such cases, the capacity of MFRL to further reduce the flight time is constrained.

\begin{figure*}[!h]
\includegraphics[width=\textwidth,trim=0.cm 0.5cm 0.cm 0.cm,clip]{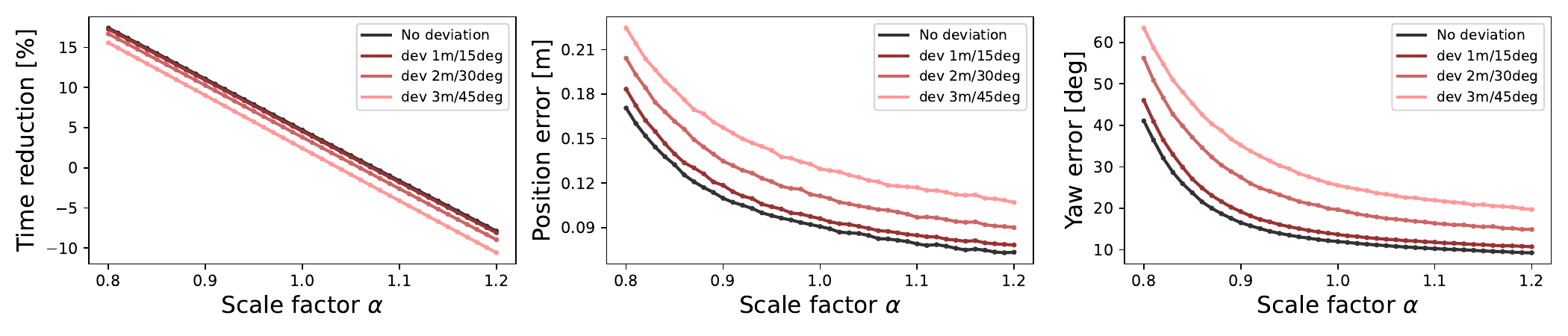}
% \vspace{-1\baselineskip}
\caption{Changes of time reduction and tracking error regarding the scale factor $\alpha$. As $\alpha$ increases, the speed profile decreases, resulting in reduced time reduction but improved tracking accuracy. Similar trends are observed when waypoints deviate randomly.}
\label{fig:sim_res_alphacomp}
% \vspace{-1.0em}
\end{figure*}

The tracking error and time reduction of the trained policy can be fine-tuned using the approach outlined in Section \ref{subsec:alg_scaling}.
Figure \ref{fig:sim_res_alphacomp} illustrates the trade-off between tracking error and time reduction based on the scale factor $\alpha$. 
When we scale down the time allocation using $\alpha$, the trajectory accelerates, but tracking error increases. 
When waypoints deviate, both the average tracking error and time reduction may adapt to these deviations, while the overall relationship between tracking error and time reduction remains consistent.

\begin{figure*}[h]
    \centering
    % \captionsetup{justification=centering}
    \includegraphics[width=0.7\textwidth,trim=0.cm 0.2cm 0.cm 0.cm,clip]{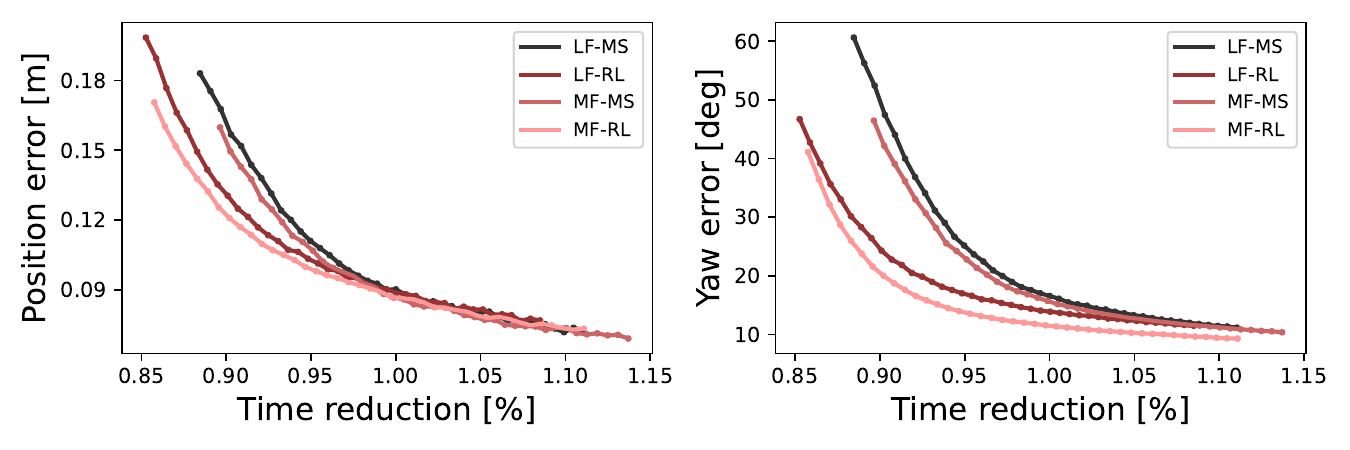}
    \vspace{-.5\baselineskip}
    \caption{Comparison of policies trained with a multi-fidelity model: the pretrained policy (\textit{MF-MS}) and the MFRL policy (\textit{MF-RL}). Additionally, comparisons include the pretrained policy (\textit{LF-MS}) and RL policy (\textit{LF-RL}) that are trained only with the low-fidelity model. The y-axis represents the relative time reduction compared to the unscaled trajectory output from the pretrained model, which is trained using multi-fidelity evaluation.}
    \label{fig:sim_res_alphacomp_time}
\end{figure*}

Utilizing this transformation, we analyze the performance improvements achieved through multi-fidelity evaluations and RL. 
By scaling the time allocation with the scale factor, we can observe trends between the scaled trajectory time and tracking error. 
Figure \ref{fig:sim_res_alphacomp_time} compares these trends between the pretrained policy and the MFRL policy. 
Additionally, we train the same policy model using only low-fidelity evaluations and compare these results with those from the pretrained and RL policies developed under low-fidelity conditions. 
Fine-tuning the model with RL shifts this trend leftward, indicating that the planning can generate faster trajectories with similar tracking errors. 
Employing multi-fidelity evaluations further enhances these trends, offering improvements over training exclusively with low-fidelity models.

\subsection{Real-World Experiment}

\begin{figure*}[!h]
\centering
    \includegraphics[width=0.98\textwidth,trim=0.cm 0.cm 0.cm 0.cm,clip]{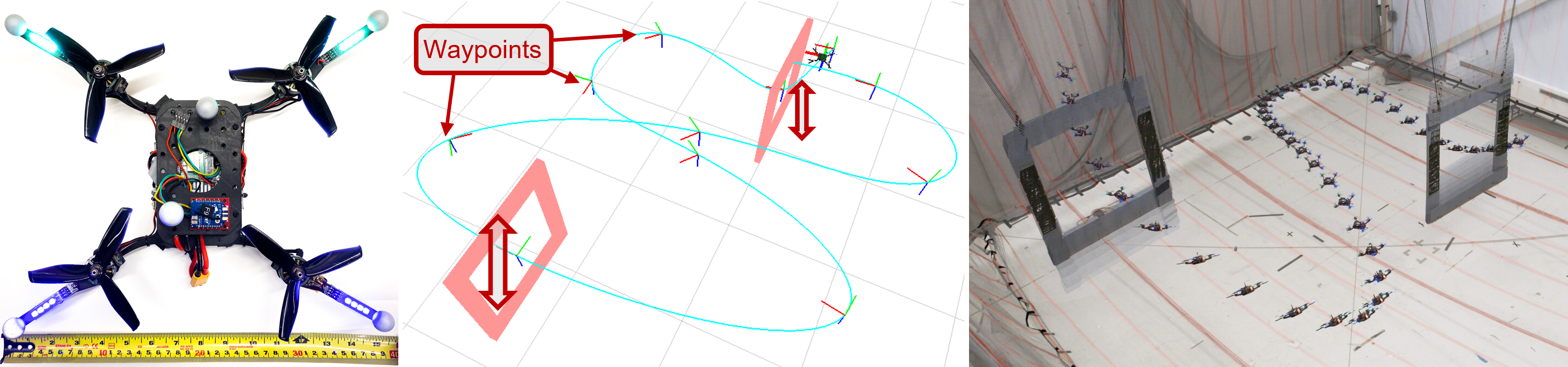}
    % \vspace{-1\baselineskip}
    \caption{Image of the quadrotor vehicle and the experimental environment utilized in the flight experiment. The trained model is evaluated by deploying it on real-world applications in which the waypoints change during flight and the trajectory must be adapted accordingly.}
    \label{fig:real_world_exp}
\end{figure*}

\begin{table*}[h!]
% \scriptsize
\footnotesize
\centering
\caption{Comparison of tracking error and trajectory time among minimum-snap trajectories (\textit{MinSnap}), trajectories generated by the policy trained in simulation (\textit{MFRL (simulation)}), and trajectories generated by the policy trained with real-world experiments (\textit{MFRL (real-world)}). \textit{$\Delta$ Time} represents the trajectory time reduction relative to the minimum-snap trajectories. The policy trained in simulation generates trajectories that are 1.69 \% faster than the baseline when waypoints are unchanged. Incorporating real-world experiment data into the training further improves the performance, resulting in a \textbf{4.26 \%} time reduction compared to the baseline.}
% \vspace{-0.5em}
\label{tab:exp_alg_res_real}
\begin{tabular}{c cc ccc ccc}
\toprule
\multicolumn{1}{c}{Deviation} & \multicolumn{2}{c}{MinSnap} & \multicolumn{3}{c}{MFRL (simulation)} & \multicolumn{3}{c}{MFRL (real-world)} \\ 
\cmidrule(lr){1-1}\cmidrule(lr){2-3}\cmidrule(lr){4-6}\cmidrule(lr){7-9}
\multicolumn{1}{c}{Pos / Yaw} &
\multicolumn{1}{c}{Pos error} &
\multicolumn{1}{c}{Yaw error} &
\multicolumn{1}{c}{Pos error} &
\multicolumn{1}{c}{Yaw error} &
\multicolumn{1}{c}{$\Delta$ Time} &
\multicolumn{1}{c}{Pos error} &
\multicolumn{1}{c}{Yaw error} &
\multicolumn{1}{c}{$\Delta$ Time}\\ 
\midrule
0.0~$\si{\metre}$ /\hfill\hspace{0.2em} 0~$\si{\degree}$  & 
0.14 $\pm$ 0.03~$\si{\metre}$ & 13.3 $\pm$ 1.5~$\si{\degree}$  & 
0.12 $\pm$ 0.02~$\si{\metre}$ & 11.4 $\pm$ 3.8~$\si{\degree}$  & 1.69 \% & 
0.13 $\pm$ 0.02~$\si{\metre}$ & 12.9 $\pm$ 3.7~$\si{\degree}$  & 4.26 \% \\

1.0~$\si{\metre}$ / 15~$\si{\degree}$ & 
0.12 $\pm$ 0.02~$\si{\metre}$ & 14.8 $\pm$ 9.1~$\si{\degree}$  & 
0.13 $\pm$ 0.05~$\si{\metre}$ & 14.6 $\pm$ 9.1~$\si{\degree}$  & 2.30 \% & 
0.13 $\pm$ 0.04~$\si{\metre}$ & 16.3 $\pm$ 9.5~$\si{\degree}$  & 4.48 \% \\

2.0~$\si{\metre}$ / 30~$\si{\degree}$ & 
0.15 $\pm$ 0.07~$\si{\metre}$ & 23.3 $\pm$ 20.2~$\si{\degree}$ & 
0.12 $\pm$ 0.02~$\si{\metre}$ & 15.6 $\pm$ 13.0~$\si{\degree}$ & 2.68 \% & 
0.13 $\pm$ 0.02~$\si{\metre}$ & 14.7 $\pm$ 7.1~$\si{\degree}$  & 3.85 \% \\

3.0~$\si{\metre}$ / 45~$\si{\degree}$ & 
0.19 $\pm$ 0.16~$\si{\metre}$ & 29.3 $\pm$ 25.9~$\si{\degree}$ & 
0.14 $\pm$ 0.05~$\si{\metre}$ & 19.4 $\pm$ 18.2~$\si{\degree}$ & 1.50 \% & 
0.13 $\pm$ 0.04~$\si{\metre}$ & 17.7 $\pm$ 16.8~$\si{\degree}$ & 2.18 \% \\
\hline
\end{tabular}
% \vspace{-0.5em}
\end{table*}

\begin{table}[h!]
% \tiny
\footnotesize
\centering
\caption{
Comparison of parameters including speed ($v$), acceleration ($a$), and motor utilization factor ($U_{\text{motor}}$) between minimum-snap trajectories (\textit{MinSnap}) trajectories generated by the policy trained in simulation (\textit{MFRL (sim)}), and trajectories generated by the policy trained with real-world experiments (\textit{MFRL (real)}) ($L_{\text{space}}=[9\si{\metre}, 9\si{\metre}, 3\si{\metre}]$).}
% \vspace{-0.5em}
\label{tab:exp_alg_res_real_spdacc}
\begin{tabular}{l ccccc}
\toprule
\multicolumn{1}{c}{} &
\multicolumn{1}{c}{$v_{\text{avg}}$} &
\multicolumn{1}{c}{$v_{\text{max}}$} &
\multicolumn{1}{c}{$a_{\text{avg}}$} &
\multicolumn{1}{c}{$a_{\text{max}}$} &
\multicolumn{1}{c}{$U_{\text{motor}}$} \\ 
\midrule

MinSnap & 3.9 \si{m/s} & 10.7 \si{m/s} & 0.8 \text{g} & 2.1 \text{g} & 15.9 $\pm$ 2.7 \% \\
MFRL (sim) & 3.7 \si{m/s} & 10.9 \si{m/s} & 0.8 \text{g} & 2.3 \text{g} & 16.1 $\pm$ 2.0 \% \\
MFRL (real) & 3.8 \si{m/s} & 11.0 \si{m/s} & 0.8 \text{g} & 2.2 \text{g} & 18.2 $\pm$ 2.6 \% \\

\hline
\end{tabular}
% \vspace{-0.5em}
\end{table}

% Explain more
We further evaluate the proposed algorithm by training the planning policy in a real-world scenario and testing the policy on a dataset of 50 randomly generated waypoint sequences, each with trajectory times $T^{\text{MS}}$ determined from real-world flight experiments. 
The reward estimator is trained using three different fidelity levels, including real-world flight experiments. 
Table \ref{tab:exp_alg_res_real} presents a comparison of the tracking error and the time reduction ($\Delta$ Time) achieved by policy models trained in simulated environments (using two fidelity levels) and real-world scenarios (utilizing three fidelity levels). 
The position and yaw tracking errors, $\bar{r}^\text{err}$ and $\bar{\psi}^\text{err}$, are calculated from the reference pose and the motion capture system measurements obtained while executing the trajectory with actual aircraft:
\begin{align}
\bar{r}^\text{err} &= \max_t\:\lVert p_r(t) - r_\text{mocap}(t) \rVert, \nonumber \\
\bar{\psi}^\text{err} &= \max_t\:|p_\psi(t) - \psi_\text{mocap}(t)|
\end{align}
where $r(t)$ and $\psi(t)$ are the reference position and yaw, respectively, and $r_\text{mocap}(t)$ and $\psi_\text{mocap}(t)$ represent the position and yaw measured using the motion capture system, as defined in \eqref{eqn:feasibility_tracking_err_3}. 
Notably, when waypoints deviate, both the models trained with MFRL in simulation and real-world experiments adapt the trajectory better than the baseline method. 
The model trained with real-world experiments further optimizes the trajectory to make it even faster than the policy only trained in simulation.
Table \ref{tab:exp_alg_res_real_spdacc} compares speed, acceleration, and motor utilization between the baseline trajectories and the trajectories from the trained policies.
Similar to the simulated results, while average speeds and accelerations are comparable, MFRL trajectories show higher maximums and a significantly greater average motor utilization.

\begin{table*}[h!]
% \tiny
\footnotesize
\centering
\caption{Comparison of computation time between the minimum-snap method (\textit{MinSnap}) and the trained policy (\textit{MFRL}). 
\textit{NLP} represents the computation time of non-linear programming, which determines the ratio between time allocations. 
\textit{Line search ($\mathcal P_1$)}, \textit{Line search ($\mathcal P_2$)}, and \textit{Line search ($\mathcal P_3$)} denote the time for determining the trajectory time with line search using the ideal dynamics model, simulations, and real-world flight experiments, respectively. 
The computation time increases by an order of magnitude between the different fidelity levels. 
\textit{QP} refers to the time for quadratic programming to determine the polynomial coefficients. 
The trained policy replaces the \textit{NLP} and \textit{Line search} procedures with neural network inference. 
It can determine the flight time in 9 \si{ms} with a PyTorch implementation (\textit{PyTorch}), and once the model is compiled with TensorRT, the inference time drops below 2 \si{ms} (\textit{TensorRT}).}
% \vspace{-0.5em}
\label{tab:exp_alg_res_comptime}
\begin{tabular}{ccccccc}
\toprule
\multicolumn{5}{c}{MinSnap} & \multicolumn{2}{c}{MFRL} \\ 
\cmidrule(lr){1-5}\cmidrule(lr){6-7}
\multicolumn{1}{c}{NLP} &
\multicolumn{1}{c}{Line search ($\mathcal P_1$)} &
\multicolumn{1}{c}{Line search ($\mathcal P_2$)} &
\multicolumn{1}{c}{Line search ($\mathcal P_3$)} &
\multicolumn{1}{c}{QP} &
\multicolumn{1}{c}{PyTorch} &
\multicolumn{1}{c}{TensorRT} \\ 
\midrule
370.9~$\si{ms}$ & 28.75~$\si{ms}$ &	33.58~$\si{s}$ & $\sim$15~$\si{min}$ & 1.72~$\si{ms}$ & 8.84~$\si{ms}$ & \textbf{1.66~$\si{ms}$} \\
\hline
\end{tabular}
% \vspace{-0.5em}
\end{table*}

\begin{figure}[!h]
\centering
\includegraphics[width=0.48\textwidth,trim=0.cm 0.2cm 0.cm 0.cm,clip]{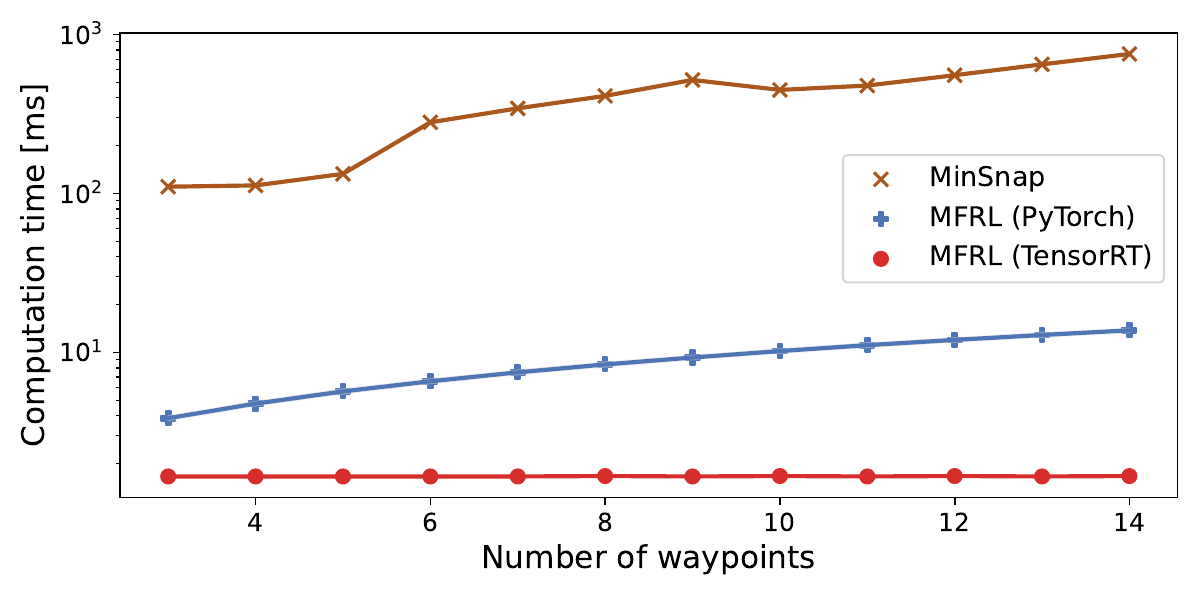}
\vspace{-0.5em}
\caption{
Comparison of computation time with respect to the number of waypoints. 
The computation time of \textit{MinSnap} increases as the number of waypoints increases, which also raises the dimension of optimization variables. 
Similarly, the computation time of \textit{MFRL} rises with the increase in the number of GRU inferences, depending on the number of waypoints. 
However, the use of the MFRL policy improves computation time by an order of magnitude.
}
\label{fig:exp_comptime}
\end{figure}

Additionally, we assess the trained model's performance in an online planning scenario, where dynamically shifting waypoints necessitate real-time trajectory adaptation within the environment shown in Figure \ref{fig:real_world_exp}. 
We put the waypoints in the middle of two gates and continuously moving the gates to update the position of the waypoints.
We commit a portion of the first trajectory segment, taking into account inference and communication delays, while adjusting the time allocation for the remainder. 
The time allocation between waypoints is determined by the trained policy, and we obtain the actual polynomial trajectory through quadratic programming. 
The trained policy is executed on the Titan Xp GPU in the host computer and communicates with the vehicle's microcontroller to update the trajectory.
By compiling the model with TensorRT, we significantly enhance the inference speed. 
This enables our trained model to adjust the trajectory in under 2 milliseconds, allowing real-time trajectory adaptation—a marked contrast to the baseline minimum-snap method, which takes several minutes for trajectory generation.
Since the planning policy comprises simple MLPs, the inference time remains consistent even on embedded GPUs, averaging 3 milliseconds when executed on Jetson Orin.
Table \ref{tab:exp_alg_res_comptime} presents a comparison of inference times between the baseline method and our trained model. 
The baseline method's line search procedure significantly increases computation time, especially when using real-world flight experiment data where evaluating multiple speed profiles for a trajectory can take several minutes. 
Generally, the minimum snap method is used with the lowest fidelity line search, but even this approach requires around 400 ms, substantially longer than our method.
Our proposed method streamlines this by substituting the line search with policy inference, making it suitable for real-time applications.
All computation times are measured on an Intel Xeon CPU with 28 cores in the host computer, except for the policy model, which is evaluated on the aforementioned Titan Xp GPU in the same machine.
Figure \ref{fig:exp_comptime} compares the inference times based on the number of waypoints. 
Even with a small number of waypoints, the policy inference time is substantially lower, requiring far less computation time.
The supplementary video includes demonstrations of the real-world experimental results.

\section{Conclusion} \label{sections:conclusion}
% Contributions
\subsection{Contributions}
We introduce a novel sequence-to-sequence policy for generating an optimal trajectory in online planning scenarios, as well as a reinforcement learning algorithm that efficiently trains this policy by combining the evaluation from multiple sources.
Our approach models the feasibility boundary entirely from data by training a reward estimator, enabling full utilization of the vehicle's capacity. 
Furthermore, we employ multi-fidelity Bayesian optimization to train the reward estimation module, efficiently incorporating real-world experiments.
This approach enables extensive policy optimization in regions where simulation fails to capture real-world phenomena.
The trained policy model generates faster trajectories with decent reliability compared to the baseline minimum snap method, even when the waypoints are randomly deviated and the baseline method fails.

\subsection{Limitations and Future Work}
The main drawback of the proposed method is the absence of a performance guarantee. 
While the trained model generates faster and more feasible trajectories for most waypoint sequences, it occasionally fails. 
We expect this limitation could be addressed by leveraging the reward estimator beyond its primary role in training. 
The trained reward estimator can potentially predict output performance before physical testing, offering crucial guarantees for high-stake dynamic systems. 
This predictive capability could ensure robustness and safety in planners, and facilitate reachability analysis for safety-critical fields.

The long training time can be improved by redesigning the reward estimator's architecture. 
We've observed that reinforcement learning convergence is influenced by the estimator's accuracy. 
In this work, we use Gaussian process for feasibility prediction due to their effectiveness with limited samples. 
As data accumulates, however, a basic neural network becomes more suitable, offering better accuracy with larger datasets. 
A challenge arises when low-fidelity datasets reach this transition threshold earlier than high-fidelity ones, forcing continued use of Gaussian process and limiting low-fidelity sample inclusion. 
Developing a model that can transition between architectures and account for varying dataset sizes across fidelity levels could lead to more efficient large-scale multi-fidelity optimization, ultimately reducing training time.

Building upon the idea of adaptive architectures, increasing the reward model's complexity and capacity could further extend the proposed method's applicability. 
Since our approach treats the full system as a black box and optimizes tests based on specific goals, a more sophisticated model could incorporate additional system parameters. 
This enhanced capacity would allow the method to handle more complex elements, such as internal controller states, or integrate with local planners like MPC. 
Currently, this work only uses IMU and motion capture inputs, which could limit sources of perception uncertainty.
A larger model capacity can ultimately extend this approach to full autonomy pipelines with various sensory inputs including camera and LiDAR, handling unique and high-dimensional uncertainties from the full system.

% \newpage

\begin{acks}
This work was supported in part by the Army Research Office through grant W911NF1910322 and the Hyundai Motor Company.
\end{acks}

\bibliographystyle{SageH}
\bibliography{refs}

\end{document}